\documentclass[libertine,twocolumn]{article}

\usepackage{geometry}
\usepackage[utf8]{inputenc} %
\usepackage[T1]{fontenc}    %
\usepackage[colorlinks]{hyperref}
\usepackage{xurl}           %
\usepackage{booktabs}       %
\usepackage{amsfonts}       %
\usepackage{graphicx}
\usepackage{color}
\usepackage{amssymb}
\usepackage{amsmath}
\usepackage{xspace}
\usepackage{nicefrac}       %
\usepackage{microtype}      %
\usepackage{xcolor}         %
\usepackage{ulem}
\usepackage{multirow}
\usepackage{soul}
\usepackage{colortbl}
\usepackage{CJKutf8}        %
\usepackage{multicol}

\usepackage[oldstyle,semibold,type1]{libertine}%
\usepackage{textcomp}%
\usepackage{mathtools,amssymb,amsthm} %
\usepackage[libertine,cmintegrals,cmbraces,vvarbb]{newtxmath}
\usepackage[scr=boondoxo]{mathalfa}%
\usepackage{bm}%

\title{The 2023 Video Similarity Dataset and Challenge}

\newcommand{\makecomment}[3]{\textcolor{#2}{[\textbf{#1}: #3]}}
\newcommand{\TODO}{{\textcolor{magenta}{\textbf{TODO}}}}

\newcommand{\vcd}{VCD\xspace}
\newcommand{\vcl}{VCL\xspace}
\newcommand{\lvcd}{\vcl}  %

\newcommand{\giorgost}[1]{\makecomment{Giorgos T}{violet}{#1}}
\newcommand{\giorgoskz}[1]{\makecomment{Giorgos KZ}{blue!20!red}{#1}}

\newcommand{\ed}[1]{\makecomment{Ed}{brown}{#1}}

\newcommand{\OURD}{DVSC23\@\xspace}
\newcommand{\DISC}{DISC21\@\xspace}

\author{
Ed Pizzi\textsuperscript{1}
\and
Giorgos Kordopatis-Zilos\textsuperscript{2}
\and
Hiral Patel\textsuperscript{1}
\and
Gheorghe Postelnicu\textsuperscript{1}
\and
Sugosh Nagavara Ravindra\textsuperscript{1}
\and
Akshay Gupta\textsuperscript{1}
\and
Symeon Papadopoulos\textsuperscript{3}
\and
Giorgos Tolias\textsuperscript{2}
\and
Matthijs Douze\textsuperscript{1}
\\ 
\\
\scalebox{0.8}{
\begin{tabular}{l}
\textsuperscript{1}Meta AI, San Francisco, Menlo Park, London, Paris \\
\textsuperscript{2}VRG, FEE, Czech Technical University in Prague\\
\textsuperscript{3}Centre for Research \& Technology Hellas, Thessaloniki\\
\end{tabular}}
}
\date{}

\newcommand{\uAP}{$\mu AP$\xspace}

\def\eg{\textit{e.g.~}}
\def\ie{\textit{i.e.~}}

\begin{document}
\maketitle

\begin{abstract}
This work introduces a dataset, benchmark, and challenge for the problem of video copy detection and localization.
The problem comprises two distinct but related tasks: 
determining whether a query video shares content with a reference video (``detection''), and additionally temporally localizing the shared content within each video (``localization'').
The benchmark is designed to evaluate methods on these two tasks, %
and simulates a realistic needle-in-haystack setting, where the majority of both query and reference videos are ``distractors'' containing no copied content.
We propose a metric that reflects both detection and localization accuracy.
The associated challenge consists of two corresponding tracks, each with restrictions that reflect real-world settings. 
We provide implementation code for evaluation and baselines\footnote{\url{https://github.com/facebookresearch/vsc2022}}.
We also analyze the results and methods of the top submissions to the challenge. 
The dataset, baseline methods and evaluation code is publicly available and will be discussed at a dedicated CVPR'23 workshop.

\end{abstract}

\section{Introduction}

Copy detection is the task of determining whether two pieces of media, \eg photos or videos, are \textit{derived from the same original source}.
This is an important capability for online services where users upload and share media, with applications including content moderation, protecting copyright, and identifying misattributed or misleading uses of media.

Video copy detection is increasingly important, as users spend more time engaging with video content online.
While video copy detection often benefits from improvements from image copy detection \cite{caron2020unsupervised,he2022vcsl}, %
video introduces unique challenges, such as localizing copied segments within a video, that requires dedicated benchmarking.

Video copy detection can be approached using any of the modalities that video contains, including visual content, audio, or even text (\eg closed captions).
In practice, distinct systems may be required for each modality, since it is useful to detect visual-only copies (\eg containing graphic content) and audio-only copies (\eg music).
In this work, we consider video copy detection as a computer vision problem, focusing on copied visual content within a video.
We concentrate on the partial video copy detection regime, where copies may contain only a fraction of the original video's footage.
Our work contributes to better understanding and advancing practical video copy detection methods.

Reporting video copies is more useful in practice if properly localized. For instance, a content moderator can
more easily determine whether a video contains harmful
content if a copy detection system can provide a time range
where a copy of harmful content is identified.
Temporal localization, simply referred to as localization
in the rest of the manuscript, is the task of identifying the
overlapping time ranges within a pair of videos, \ie predicting copied ``segments''.

\newcommand{\imcreds}[2]{\raisebox{\depth}{\scalebox{0.4}{\rotatebox{270}{#2}}}}

\begin{figure}[t]
    \centering
    \includegraphics[width=0.46\linewidth]{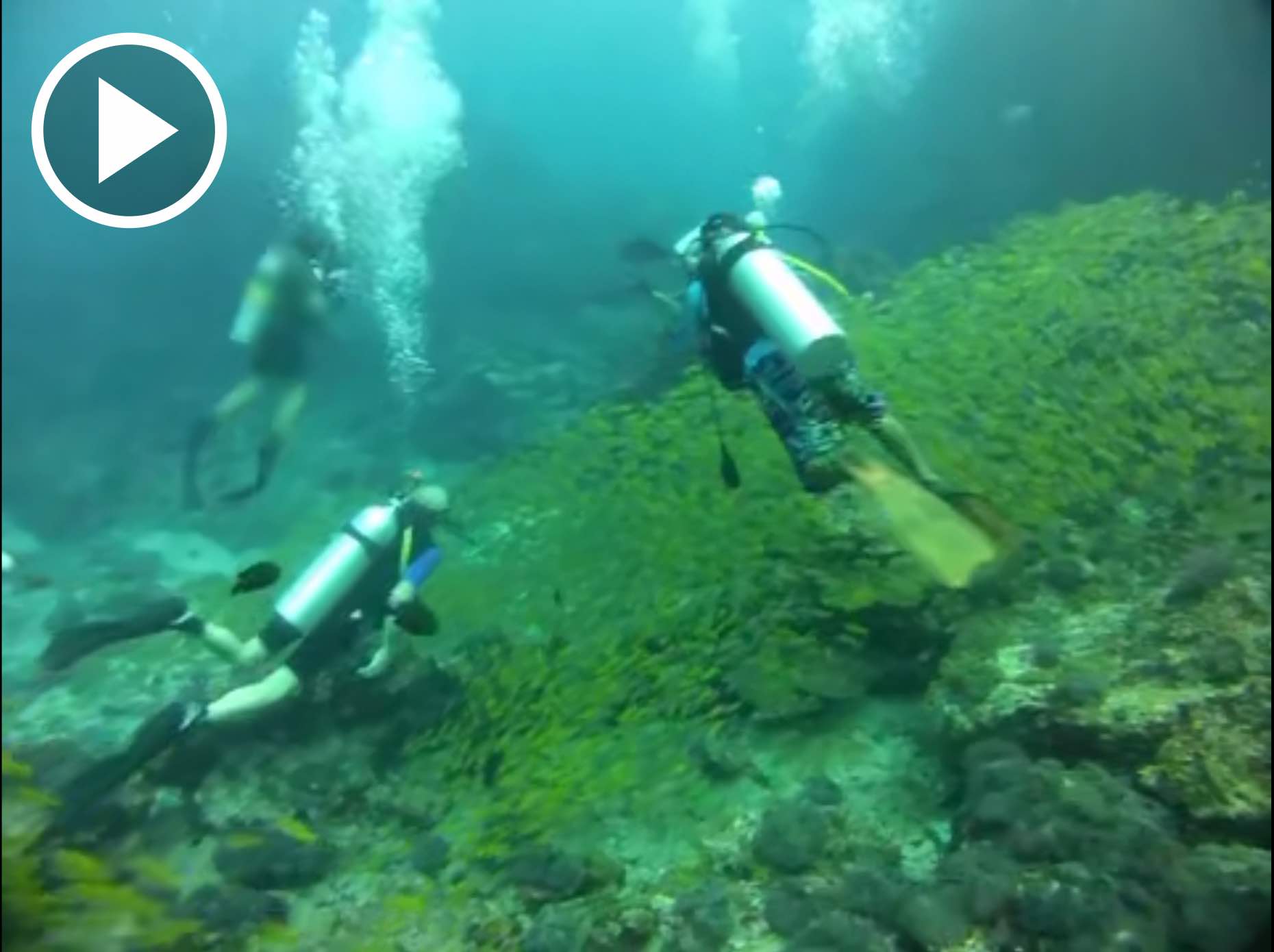}
    \imcreds{R215026}{comicpie}
    \includegraphics[width=0.46\linewidth]{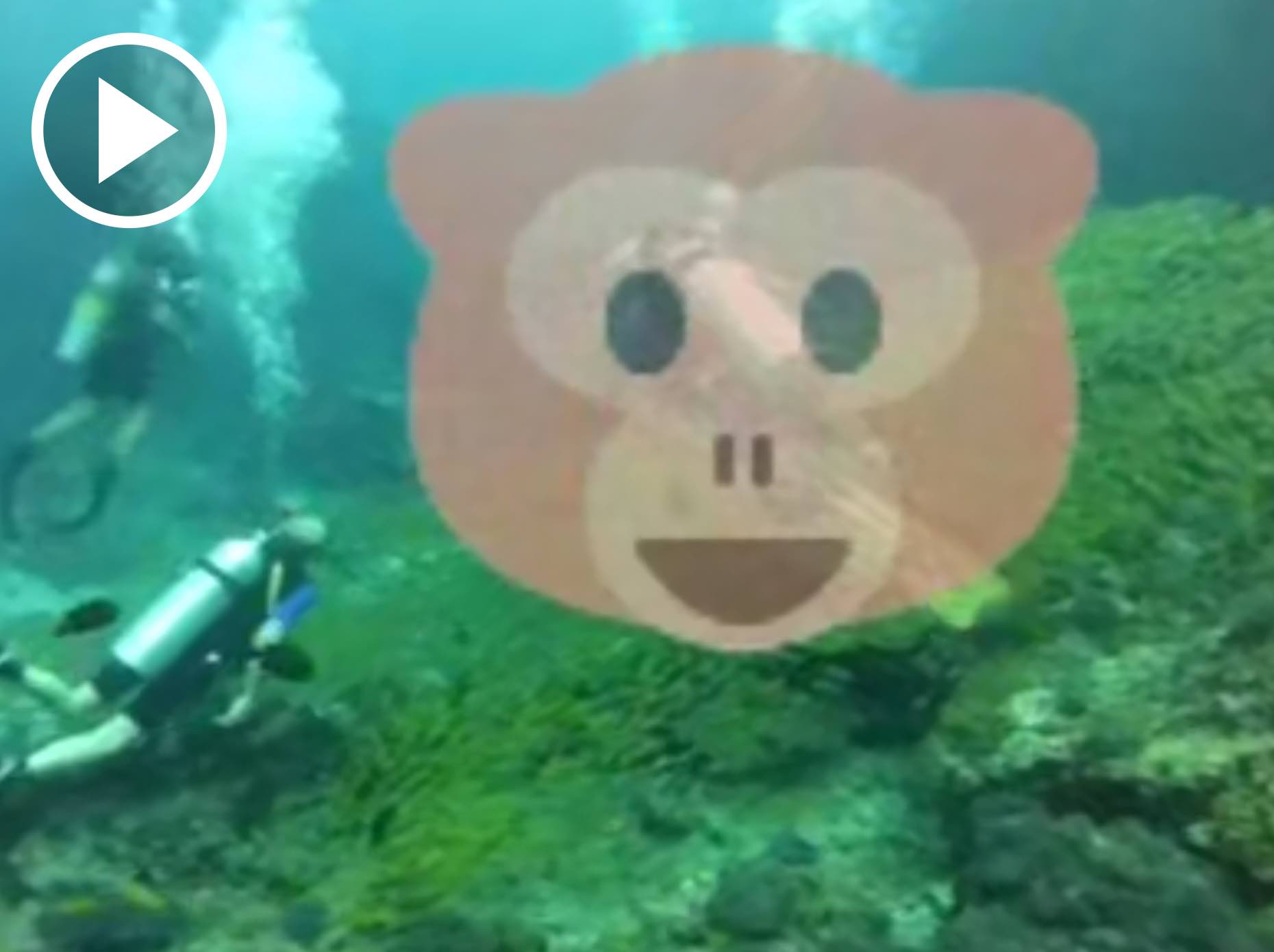} 
    \imcreds{Q201671}{comicpie, chrishusein}
    \caption{Example of a frame of the original video (left, denoted by ``reference video'' in this work) and a frame of its edited copy (right, denoted by ``query'' video).
    }
    \label{fig:intro}
\end{figure}

\subsection{Use cases}

Video copy detection systems are used to moderate content at large scale, enabling automated systems to take down modified copies of previously flagged media~\cite{douze2021isc,FBAI2020misinfo}.
This is particularly important for problem classes such as misinformation, which can be difficult for automated systems, or even non-expert humans, to classify~\cite{FBAI2020misinfo}.
In these cases content can become
``viral'', \ie it spreads out at an exponential pace. 
Video copy detection allows taking action on copies of harmful content more quickly than human moderation or user reports.
In addition, copy detection systems can reduce the harmful content that both users and content moderators are exposed to.

Video copy detection helps identifying the original source of a video, for example to determine when a video is misattributed, manipulated or shared in a misleading way.
It is also used to protect copyright, \eg by identifying copies of movies uploaded to video sharing platforms or identifying content of videojournalists used without a license, and is a strong form of deduplication.
Deduplicating media shown to users, \eg in recommendations or search results, can improve diversity and reduce repetition.
Deduplication also improves machine learning datasets by reducing memorization and overfitting~\cite{dalle2022pretraining,somepalli2022diffusion}.

\subsection{Challenges}

\paragraph{Scale.}

Video copy detection is challenging in practice due to the large scale at which real-world systems operate.
Very large database sizes increase the risk of false-positive detections, which is a challenge given the high precision required by many use cases.
The large amount of information contained within each video poses engineering challenges at scale, making the efficiency of such systems important.

\paragraph{Transformations.}

Videos may be significantly transformed.
Users may edit videos for aesthetic or artistic reasons, such as cropping a video or applying an Instagram filter.
Users may make modifications that have a misleading or misinforming effect, in which case matching to the original source can provide important context.
The modifications are not necessarily intended by the the user. 
For example, screen-casting often captures additional user-interface elements from the user's screen, and video sharing platforms typically resize or re-encode videos for storage or distribution.
However, videos may be edited specifically to evade content moderation.
A system needs to detect and localize multiple copied segments for a pair of videos, 
even for segments that constitute only a small fraction of the total duration of one or both videos.

%

\subsection{Problem definition}

This benchmark considers the problem of video copy detection and localization, where a large database of \textit{reference} videos are searched using a separate set of \textit{query} videos. %
We specifically focus on \textit{partial} video copy settings, where copied segments may be contained within larger videos.
We study video copy detection as a computer vision task, focusing on the visual content of videos.
To the extent possible within the limitations of our dataset size,
we aim to replicate the conditions in which applied practical systems operate. %
To that end, the majority of both query and reference videos are \textit{distractors} that do not correspond at all.

We define two tasks that are evaluated on our dataset.

\paragraph{Video copy detection.}
The video copy detection (\vcd) task requires identification of query-reference video pairs that contain copied content, without localizing the common content within the videos.
Predictions are specified as tuples of the form
{\it(query-video, reference-video, confidence-score)},
identifying a pair of query and reference videos and a score indicating the confidence of each prediction.
It is a detection task in the sense that a set of copied videos needs to be identified, in contrast to a ranking/retrieval task where all videos are ranked according to the confidence of the copy prediction. In detection, the confidence is typically thresholded to result in the set of prediction tuples. Therefore, 
we propose to evaluate this task using \uAP, a global ranking metric, that captures performance under all possible thresholds.

\paragraph{Video copy localization.}
The video copy localization (\vcl) task requires to identify the temporal segments within a pair of videos that contain copied content.
\vcl predictions are specified as tuples of
\textit{(query-video, database-video, localization-box, confidence-score)},
where \textit{localization-box} contains the start and end timestamp for both the query and reference video.
A \vcl system may predict multiple segment predictions for a pair of query and reference videos, each with a different localization and score.
We extend the standard \uAP metric to take the localization accuracy into account. 
%
%
%
%
%
%

%
%

%
%

%
%
%
%
%
%
%

%
%
%

\section{Related work}
\label{sec:related}
\begin{table*}
    \centering
    \def\sp{\hspace{5pt}}
    \def\spp{\hspace{10pt}}
    \begin{tabular}{l@{\sp}c@{\sp}c@{\sp}c@{\sp}c@{\sp}r@{\sp}r@{\spp}r@{\spp}c@{\sp}c@{\sp}c@{\sp}}
        \hline
        \multirow{2}{*}{Dataset} & \multirow{2}{*}{Domain} & \multirow{2}{*}{Task} & \multirow{2}{*}{Relations} &  \multirow{2}{*}{Edit type} & \multicolumn{3}{c}{Dataset size} & \multirow{2}{*}{Year} & \multirow{2}{*}{Comp.} \\
        \cmidrule(rr){6-8}
         & & & & & ~Queries~ & ~~~QueriesD & Database~ & & \\
         \hline \hline
        
        Copydays~\cite{Douze2009EvaluationOG}          & Image & D   & C   & U+S & 157 & -   & 3k   & 2009 & No  \\
        \DISC~\cite{douze2021isc}                      & Image & D   & C   & U+S & 20k & 80k & 1M   & 2021 & Yes \\ \hline
        Muscle-VCD~\cite{law2007muscle}                & Video & D   & C   & U   & 18  & -   & 101  & 2007 & No  \\
        CCWeb~\cite{wu2007ccweb}    			       & Video & R   & C   & U   & 24  & -   & 13k  & 2007 & No  \\
        TrecVid~\cite{awad2014content}                 & Video & D+L & C   & S   & 11k & -   & 11k  & 2008 & Yes \\
        UQ\_Video~\cite{song2011multiple}              & Video & R   & C   & U   & 24  & -   & 167k & 2011 & No  \\ 
        EVVE~\cite{revaud2013event}                    & Video & R   & E   & U   & 620 & -   & 2.4k (+100k) & 2013 & No \\
        VCDB~\cite{jiang2014vcdb}   			       & Video & D   & C   & U   & 528 & -   & 100k & 2014 & No \\
        FIVR-200K~\cite{kordopatis2019fivr}            & Video & R   & C+I & U   & 100 & -   & 226k & 2019 & No \\ 
        SVD~\cite{jiang2019svd}                        & Video & R   & C   & U   & 206 & -   & 562k & 2019 & No \\ 
        VCSL~\cite{he2022vcsl}                         & Video & L   & C   & U   & 28k pairs & 28k pairs & - & 2022 & No \\ 
        \rowcolor{orange!70} \OURD                     & Video & D+L & C   & S   & 2k  & 6k  & 40k  & 2023  & Yes \\
        \hline
    \end{tabular}
    \caption{Datasets that are publicly available and are related to our dataset and competition. Task: Detection, Localization and Retrieval. Relations: Copy, Incident and Event. Edit type: User-generated and Synthesized transformations. Comp.: whether there was a corresponding research competition. QueriesD: distractor queries.
    \label{tab:datasets}}
\end{table*}

In this section, we provide an overview of some of the fundamental works that have contributed to the field of video copy detection and localization, as well as several related tasks and datasets.

\subsection{Tasks and metrics}

\paragraph{Detection and localization.}
Task formulations from the literature that are closely related to our tasks are VCDB~\cite{jiang2014vcdb} and VCSL~\cite{he2022vcsl}.
VCDB considers a set of trimmed query segments to detect all videos from a reference set that contain overlapping content to the queries. 
A video is considered correctly detected if it shares a least one frame with the query segment, which is evaluated with \textit{precision}, \textit{recall}, and \textit{F1} measures. 
Methods are evaluated on a single operating point, requiring the selection of a proper threshold. 
VCSL is a video copy localization task evaluated using  video pairs for the precise localization of shared content. 
\textit{Precision} and \textit{recall} are are computed from the overlap of the predicted and ground truth segments and yield a \textit{F1} score. 
This approach also evaluates methods on a single operations point. 
By contrast, we use the \uAP that synthesizes multiple thresholds for both of our tasks. For \vcd, we measure the ranking quality based on global similarity scores. Also, our queries contain irrelevant content, \ie they are not well-trimmed.
For \vcl, we rely on VCSL's definitions for \textit{precision} and \textit{recall}, and consider all query-reference video pairs for the metric calculation. 

\paragraph{Retrieval.}
Another line of research adopts the retrieval counterpart of \vcd. 
Near-duplicate video retrieval is the most prevalent formulation, as benchmarked in CCWeb~\cite{wu2007ccweb}. 
The goal is to search, in a database of reference videos, for video copies to a given query and rank them based on a global score. 
The evaluation metric is per-query average precision, which is averaged over queries into the \textit{mean Average Precision} (\textit{mAP}). 
Hence, what differentiates retrieval settings from ours is that only per-query ranking matters. %
In our setup, the ability to apply a similarity threshold to detect relevant video pairs matters.

Finally, relations between two videos can be considered similar at different granularity levels. %
In this work, we considered only video copies, \ie, edited versions of the same source video. %
In other works, the adopted definitions further consider videos of the same incident~\cite{kordopatis2019fivr}, \ie, videos capturing the same spatio-temporal span, or videos of the same event~\cite{revaud2013event}, \ie, videos capturing the same spatial or temporal span. 
These definitions are usually combined with retrieval settings and \textit{mAP} as the evaluation metric.
    
\subsection{Methods}

\paragraph{Video copy detection.} 
VCD methods can be roughly classified into two categories: \ie video-level and frame-level methods. 
Video-level methods extract a global representation for the video and then compare videos using standard (dis-)similarity measures, \eg dot product or Euclidean distance. 
These methods reduce the problem to the embedding of videos in high-dimensional spaces.    
They extract hand-crafted or deep features from all video frames and use aggregation schemes, \eg mean pooling~\cite{wu2007ccweb,kordopatis2017near}, Bag-of-Words~\cite{cai2011million,kordopatis2017nearb}, or hashing \cite{song2011multiple}, to generate global video vectors. 
Such approaches are less effective for partial copy detection, as the aggregation of features is polluted by clutter and irrelevant content. 

Frame-level methods represent videos with multiple vectors and involve fine-grained similarity functions with spatio-teporal representations. 
Several methods generate spatio-temporal representations with transformer-based networks~\cite{shao2021temporal,he2022learn}, multi-attention networks~\cite{wang2021attention}, or Fourier-based representations~\cite{Baraldi2018LAMVLT}, for temporal aggregation. 
Recent works learn parametric matching functions to estimate the video-to-video similarity~\cite{kordopatis2019visil,kordopatis2022dns,kordopatis2023self}. 
They employ a video similarity network~\cite{kordopatis2019visil} that captures fine-grained spatial and temporal structures within pairwise video similarity matrices.
The network is trained either with supervision~\cite{kordopatis2019visil}, knowledge distillation from a trained teacher~\cite{kordopatis2022dns} or seld-supervision~\cite{kordopatis2023self}.
    
\paragraph{Video copy localization.} 

Early \vcl methods rely on hand-crafted schemes for copy localization.
A straightforward approach to detect copied segments is to check the temporal consistency of matching frames using  temporal Hough voting~\cite{douze2010compact}. 
Alternatively, temporal networks~\cite{tan2009scalable} consider a graph considering the matched frames and their similarities as nodes and edges. 
The matching is then cast as a network flow optimization to extract the longest paths, which are presumably copied segments. 
Another solution using dynamic programming~\cite{chou2015pattern} so as to extract the diagonal blocks with the largest similarity allows limited horizontal and vertical movements to increase flexibility. 
Circulant Temporal Encoding aligns videos in the Fourier domain~\cite{revaud2013event,douze2016circulant} representations to avoid the complexity of pairwise frame matching.
It was generalized with temporal matching kernels~\cite{poullot2015temporal} and trainable coefficients~\cite{Baraldi2018LAMVLT}. 
Video similarity networks detect the boundaries of the copied segments based on a repurposed object detection model~\cite{jiang2021learning}, or directly predicting the probability of a frame pair lying right on the partial alignment~\cite{han2021video}.
    
\subsection{Datasets and competitions}

\paragraph{Datasets.}
    Several datasets related to our \OURD have been proposed in the literature. 
    The most relevant ones are the VCSL~\cite{he2022vcsl} and VCDB~\cite{jiang2014vcdb}, which have been compiled for video copy detection and video copy localization, respectively. 
    They consist of videos that share overlapping copied segments; hence, they contain segment-level annotations indicating the specific timestamps of the video overlaps. 
    Other related datasets are CCWeb~\cite{wu2007ccweb} and SVD~\cite{jiang2019svd} that simulate the problem of video copy detection in video collections. 
    They contain video-level annotations for the relevant videos to a given query. 
    FIVR-200K~\cite{kordopatis2019fivr} and EVVE~\cite{revaud2013event} adopt broader definitions for video associations, considering relevant videos that depict the same incident or event. 
    All the aforementioned datasets consist of user-generated videos that have been collected and manually annotated. 
    By contrast, our dataset contains synthetically generated queries based on a number of carefully selected transformations that can simulate real cases of video copies. 
    In that way, we can go into larger scales compared to previous VCL datasets and conduct a systematic evaluation of the submitted solutions. Also, we include a large number of distractor queries that make the detection and localizations tasks more challenging.

    \paragraph{Competitions.}
    A very closely related competition was the content-based copy detection track in the TrecVid series, that ran between 2008 and 2012~\cite{awad2014content}. 
    However, it was discontinued due to near perfect performance of the participant teams. 
    Therefore, our goal is to
    provide a more challenging benchmark that will attract the research community's interest. 
    Additionally, ISC2021~\cite{douze2021isc} is the sibling competition of the current challenge with a focus on the image domain.

\section{Dataset}
\label{sec:dataset}

\newcommand{\igF}[1]{\includegraphics[width=3.5cm]{figs/example_vcd/#1}}

The challenge dataset is constructed from a subset of YFCC100M videos~\cite{Thomee2016YFCC100MTN} by applying a series of filters and transformations.

\subsection{Pre-processing}

We filter YFCC100M videos to select videos that are under creative commons licenses that allow their inclusion in derivative works.
We also remove videos that were either too short (less than 5 seconds) or too low resolution (less than 320x320).

\begin{figure}
    \centering
    \begin{tabular}{c|c}
        \igF{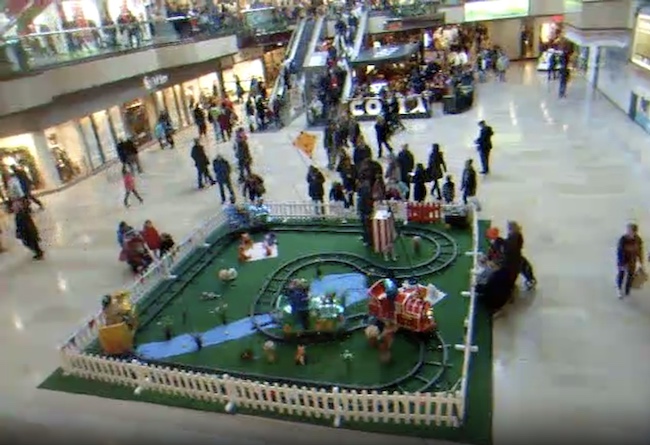} & \igF{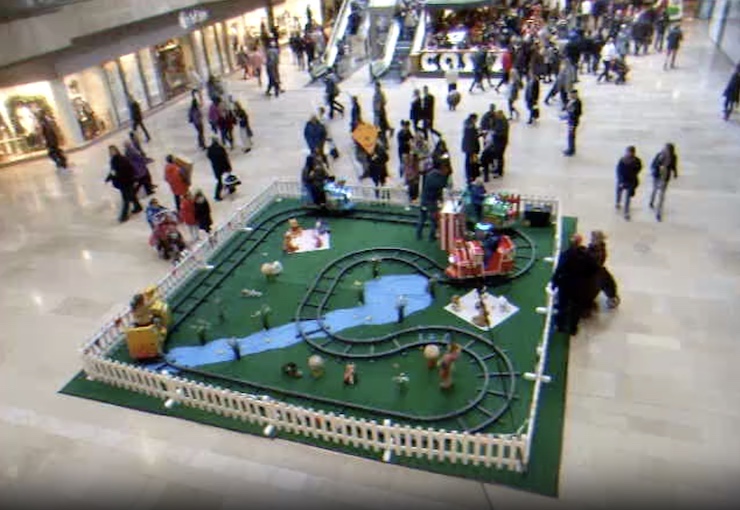} \imcreds{?}{thievingjoker}\\
        \igF{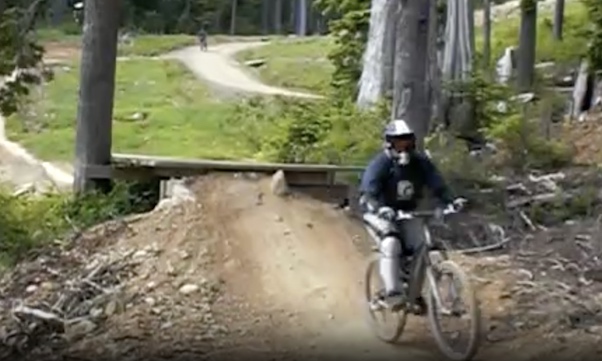} & \igF{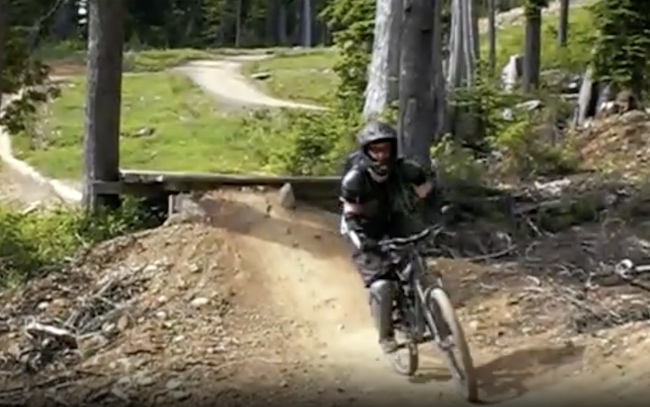} \imcreds{?}{Gwydion M. Williams}
    \end{tabular}
    \caption{Two examples (top-bottom) of frames from different (left-right), not copied, videos that are visually very similar.}
    \label{fig:false_close_duplicates}
\end{figure}

We use a semi-automated process to remove naturally occurring copies in the dataset.
We use multiple proprietary copy detection systems at high-recall / low-precision operating points, to identify potential copies, then use human labeling to distinguish true copies from challenging negatives.
These video pairs contain mostly similar scenes, however have subtle differences to indicate that they are from different videos rather than copies. 
For example, Figure~\ref{fig:false_close_duplicates} shows a bicyclist  in one video, and another bicyclist captured at the same scene in another video.
Some challenging negatives, \ie video pairs with no copied content, identified by human labeling are incorporated into the dataset.

The next step consists in applying transformations to the dataset videos to preserve people’s privacy. 
We remove the audio track of all videos to avoid inadvertent capture of sensitive conversations. 
We avoid including recognizable people. 
Unlike the images case~\cite{douze2010image}, excluding all videos with recognizable people at any point in the video results in little usable content, so it is not an option. 
Instead, we detect persons using an image segmentation model on the video frames and  implement a blurring that renders people unrecognizable. 
The blur radius is adaptive, larger for detection regions that take up a larger part of the frame.

\subsection{Dataset Structure}

We create three dataset splits: training, validation and test.
Each split contains unedited reference videos and a set of query videos, some of which contain copied sections from one or more reference video.
We additionally include 500 challenging negative pairs identified by human labeling in each split.
The test set contains only queries, searched against the validation reference set.
The composition of each dataset split can be seen in Table \ref{tab:dataset_composition}.
We replicate applied VCD conditions where most queries contain no copied segments.

\begin{table}
\centering
\scalebox{0.8}{
\begin{tabular}{lrrrrr}
    \toprule
          &         &            &   copied & queries with & distractor \\
    split & queries & references & segments &       copies &    queries \\
    \midrule
    train &    8404 &      40311 &     2708 &         1929 &       6475 \\
      val &    8295 &      40318 &     2641 &         1926 &       6369 \\
     test &    8015 &            &     2519 &         1840 &       6175 \\
    \bottomrule
\end{tabular}
}
\caption{Composition of each dataset split.}
\label{tab:dataset_composition}
\end{table}

\begin{figure}
\includegraphics[width=\columnwidth]{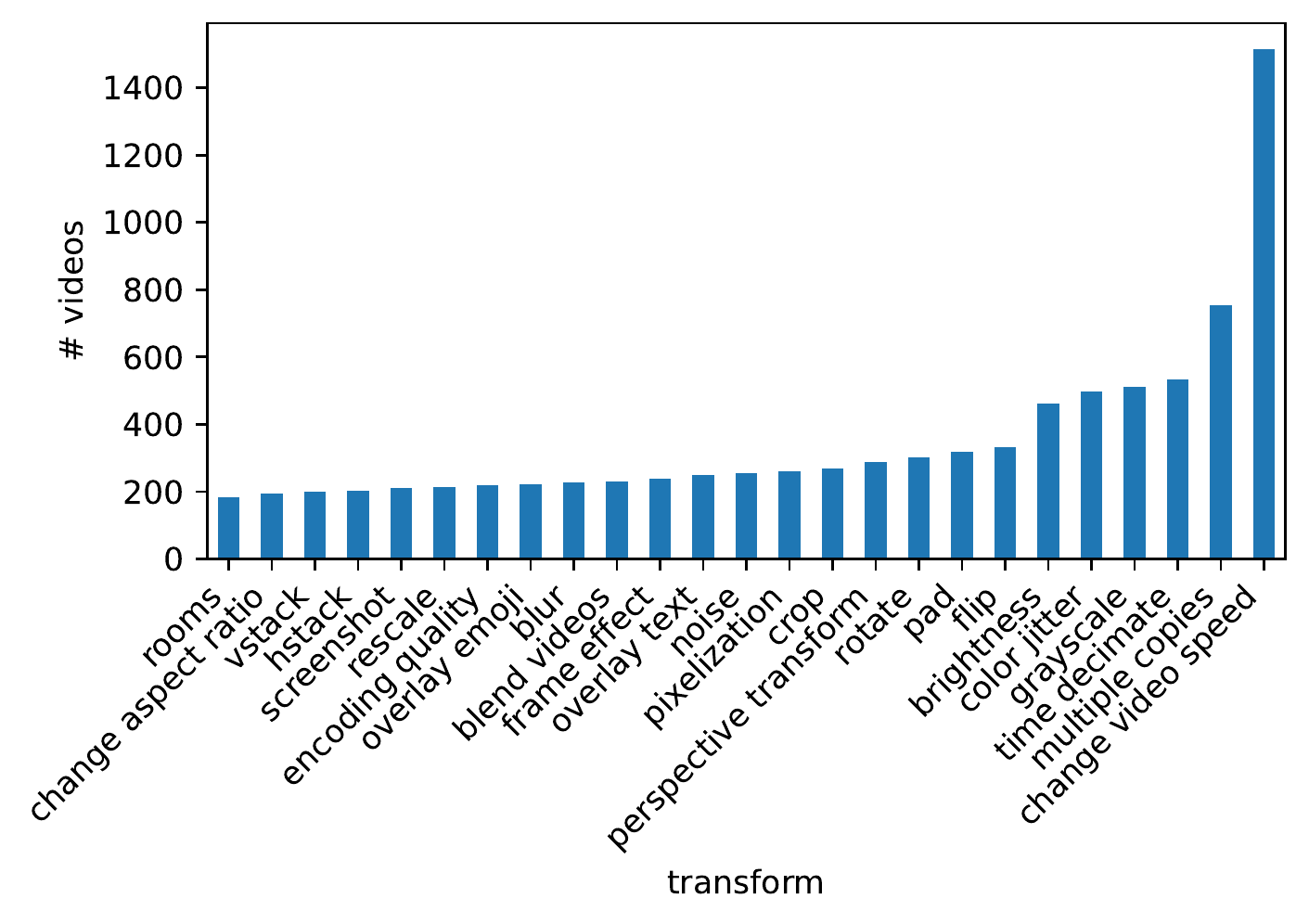}
\caption{Distribution of transforms used in the dataset.}
\label{fig:transdist}
\end{figure}

\paragraph{Differences between splits.}
We add additional transforms to validation and test splits to encourage solutions that do not overfit to the training transforms.
We also transform $50\%$ of training and validation queries so that detecting transformations is not a strong predictor of distractor queries.

\subsection{Transforms}

AugLy~\cite{augly} augmentations are used to create transformed query videos.
Queries may have one or more copied sections from the reference set that are transformed using multiple augmentations.

Queries with copied segments are created by inserting part of the reference video into the query video.
Videos are then transformed using multiple augmentations sampled from the following transforms:
\begin{itemize}
    \item Color modification - brightness, color jitter, grayscale;
    \item Add overlays - emojis, text, add frame effect;
    \item Stack videos - horizontally, vertically;
    \item Pixel modifications - blur$\dag$, noise*, pixelize*, change encoding quality, blend videos;
    \item Spatial updates - crop, add padding, rotate, flip, change aspect ratios, rescale;
    \item Temporal updates - change video speed, ``time decimate'' (retain alternating evenly-spaced segments).
\end{itemize}
Additionally there are complex transformations added for specific product scenarios: %
\begin{itemize}
    \item Overlay onto screenshot$\dag$ - simulate a screen capture of the video shared on social media
    \item Rooms* - arrange multiple videos in a grid, simulating video conference recordings
    \item Perspective transform and shake* - simulate an unsteady manual recording of a screen playing the video
\end{itemize}

\begin{figure}
\includegraphics[width=\columnwidth]{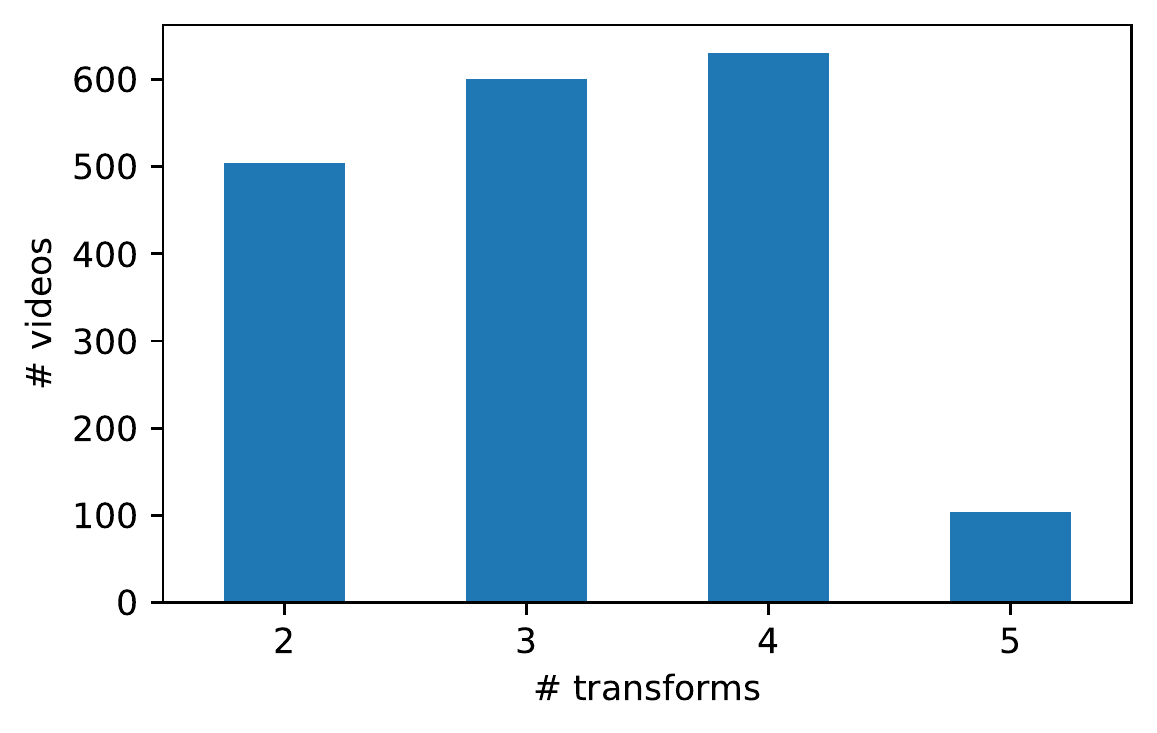}
\caption{Distribution of the number of transforms per video in the dataset.}
\label{fig:numtransdistribution}
\end{figure}

\begin{figure}
    \centering
    \begin{tabular}{cc}
    \multicolumn{2}{c}{rotate, overlay emoji} \\
    \igF{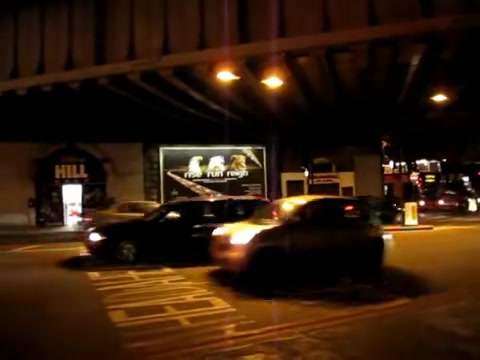} \imcreds{R206774}{erin \& camera} &
    \igF{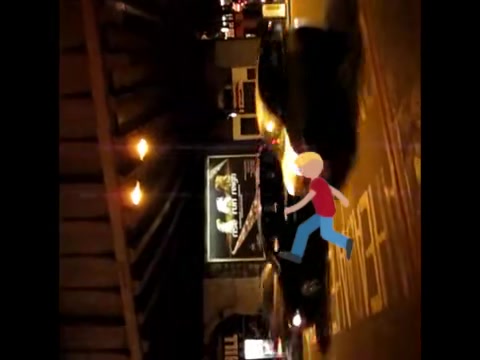} \imcreds{Q303638}{fczuardi, kylegordon, erin \& camera, aballant38} \\
    \hline
    \multicolumn{2}{c}{blend videos, change video speed} \\
    \igF{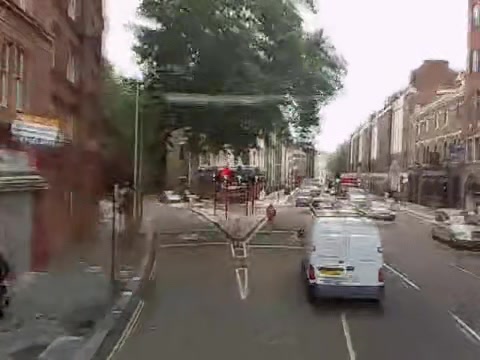} \imcreds{R218686}{fsse8info} &
    \igF{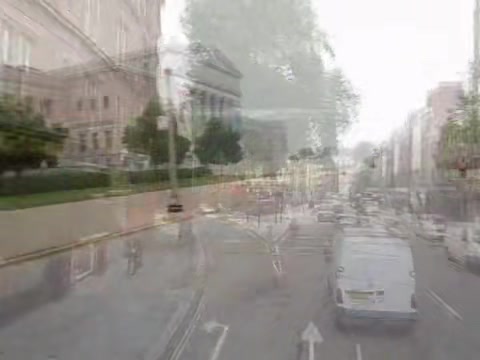} \imcreds{Q307298}{WalterPro4755, simonrobic, fsse8info, ltbluesoda} \\
    \hline
    \multicolumn{2}{c}{grayscale, overlay onto screenshot} \\    
    \igF{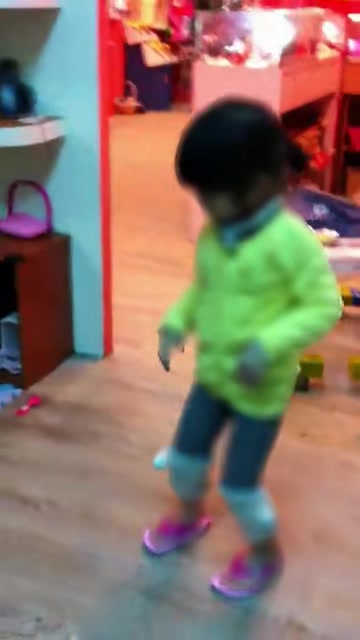} \imcreds{R211744}{
        \begin{CJK*}{UTF8}{gbsn}我愛荳芽\end{CJK*},
        \begin{CJK*}{UTF8}{gbsn}蔡\end{CJK*}
    } &
    \igF{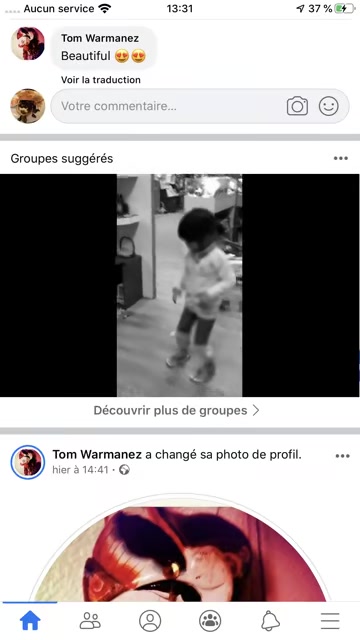} \imcreds{Q302052}{
        \begin{CJK*}{UTF8}{gbsn}我愛荳芽\end{CJK*},
        \begin{CJK*}{UTF8}{gbsn}蔡\end{CJK*},
        bibi95, Loimere, juniorbonnerphotography
    }  \\
    \hline
    \multicolumn{2}{c}{grayscale, rooms} \\
    \igF{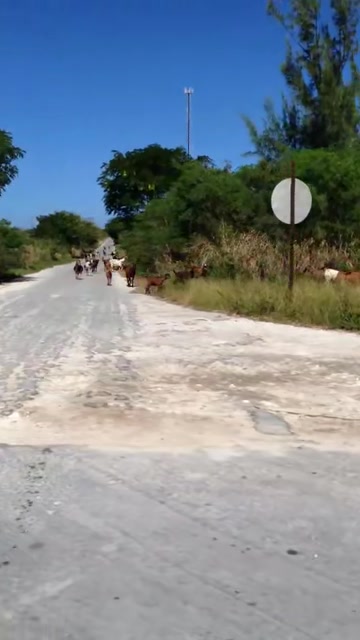} \imcreds{R231964}{denisema4} &
    \igF{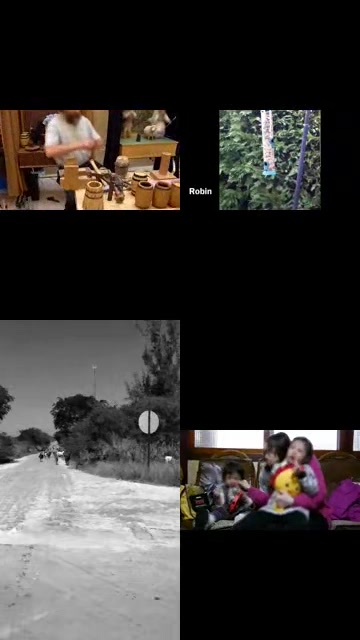} \imcreds{Q300437}{
        \begin{CJK*}{UTF8}{gbsn}arch林\end{CJK*},
        forestwildlife, cmoewes, kaleighak, jean-louis zimmermann, denisema4} \\

    \end{tabular}
    \caption{    
        Examples of copied video frames. Each copy is obtained with at least two image transformation.  
    }
    \label{fig:frame_examples}
\end{figure}

This generative recipe would typically embed one temporal segment from a reference video into a query video. 
To replicate more complex cases observed in production, there are 2 additional temporal edits: (1) multiple segments from the same reference video are added to a query and (2) multiple segments from different reference videos are embedded into a query video. 
These types of temporal edits can be observed in mashups, e.g. sports highlights.
We choose sampling parameters for each augmentations to make copied segments challenging yet recognizable.
Figure~\ref{fig:transdist} shows the number of types each transform is used in the dataset, while Figure~\ref{fig:numtransdistribution} shows how many videos are the outcome of composing 2,3,4, and 5 transforms.
Transformations marked with * are only used in validation and test sets, and transformations marked with $\dag$ are used only in the test set.
See Figure~\ref{fig:frame_examples} for example of some copied frames.

\section{Evaluation metric}
\label{sec:metric}

\label{metrics}
For both the \vcd and \vcl tasks, we consider a metric that operates on all queries jointly and takes the confidence scores into account. %
Given the ranked listed of predictions, where the ranking is performed based on the confidence in descending order, we use the area under the precision-recall curve. 
Note that a single precision-recall curve is generated jointly for all queries, and not separately per query.
This measure is also known as micro-AP (\uAP), and is used for instance-level recognition~\cite{perronnin2009family,google2019landmarkschallenge}, and for image copy detection~\cite{douze2021isc}.
It is estimated as
\vspace{-3pt}
\begin{equation}
\mu AP = \sum_{i=1}^{N} P(i) (R(i)-R(i-1)) \in [0, 1],
\label{equ:ap}
\end{equation}
where $P(i)$, and $R(i)$ are the precision and recall, respectively, at position $i$ of the sorted list, and $N$ is the total number of predictions.
Distractor queries have no positive result, so any tuple with a distractor query decreases the \uAP score. 
In the following we define how to estimate \uAP for video copy detection and localization. In particular, we describe how to estimate precision and recall for each case. 

\begin{figure*}[t]
    \centering
    \includegraphics[width=0.9\textwidth]{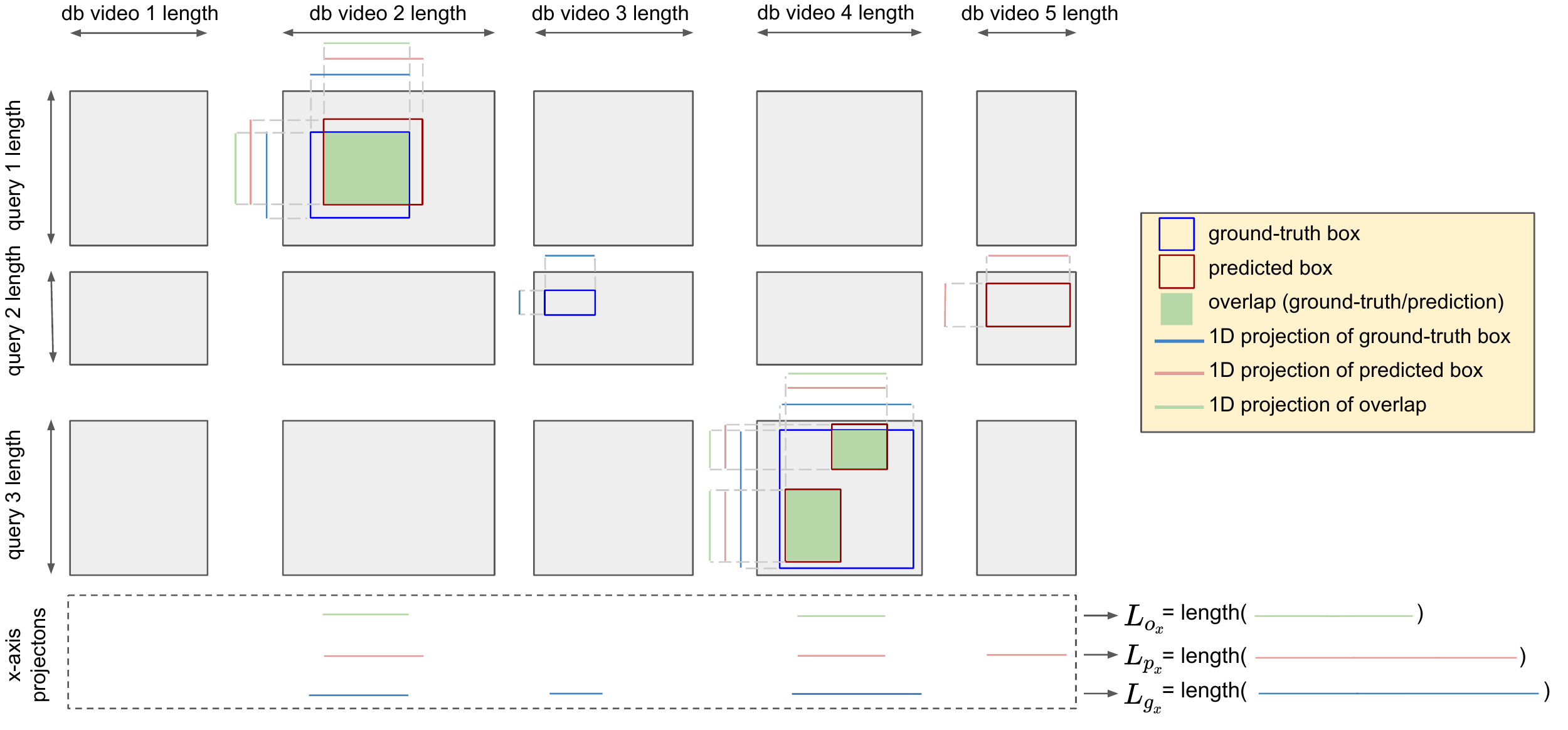}
    \caption{Toy example for the estimation of precision/recall for video copy detection with temporal localization. Example with 3 queries, 5 reference videos, 3 ground-truth boxes and 4 predicted bounding boxes. The length is estimated for the union of all line segments; the figure illustrates the x-axis projection (db videos), while the case for the y-axis (query videos) is estimated in the same way. 
    }
    \label{fig:pr_explained}
\end{figure*}
\begin{figure*}[t]
    \centering
    \includegraphics[width=0.9\textwidth]{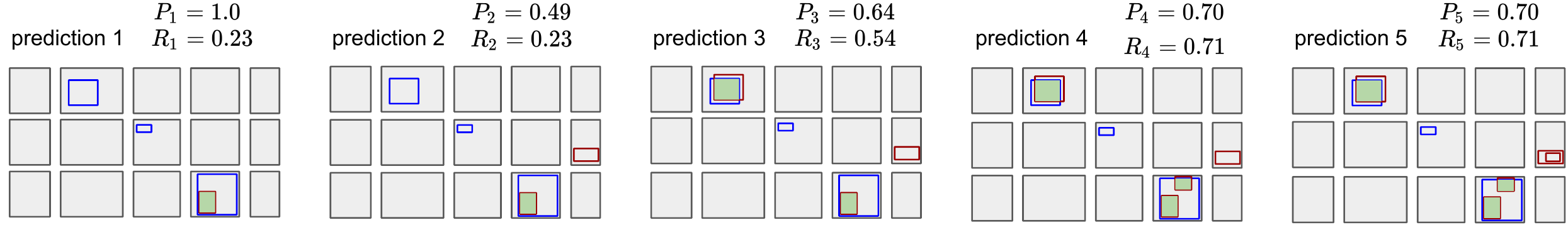}
    \caption{Estimation of precision and recall for a toy example with a list with 5 predicted bound boxes sorted from the most to the least confident. The state with the 4 predictions corresponds to the example of Figure \ref{fig:pr_explained}.
    }
    \label{fig:pr_examples}
\end{figure*}
%

%
\paragraph{\vcd metric.}
\vcd predictions include a single score per query-reference video pair, and can be evaluated with a standard \uAP metric, using the conventional definition of precision and recall.

A predicted pair is considered a correct detection if that video pair is a ground-truth match, \ie contains a copied segment.
At rank $i$, precision is simply the number of correct detections up to that rank over $i$, while recall is the number of correct detections up to that rank over the total number of ground-truth matches between all queries and all reference examples. 
This definition is identical to that used for the Image Similarity Dataset and Challenge~\cite{douze2021isc}.

\paragraph{\lvcd metric.}
In the case of a temporal localization task, we extend the definition of \uAP to take the quality of the predicted boxes into account. 
To achieve this, we adopt, with a slight modification, the evaluation metric used in the VCSL dataset~\cite{he2022vcsl}.
The ground-truth between a reference and a query video is defined as a 2D bounding box in a space where the x-axis (resp. y-axis) is the temporal dimension of the reference (resp. query) video, corresponding to a copied segment.
Similar 2D bounding boxes are used to define the predictions for localization of a method.
Then, the overlap between ground-truth and prediction are given by the intersection of boxes (also a box).
In case of multiple boxes of a kind for a video pair, we consider the projection of the union of all boxes.
At rank $i$,  which includes $i$ predicted boxes, the length of the union of these projections on the x-axis for all query and reference video pairs is denoted by $L_{g_x}(i)$, $L_{p_x}(i)$, and $L_{o_x}(i)$, for the ground-truth, predictions, and overlap, respectively. Similarly for the y-axis.
At rank $i$, precision and recall are given by 
\begin{equation}
P(i) = \sqrt{\frac{L_{o_x}(i)L_{o_y}(i)}{L_{p_x}(i)L_{p_x}(i)}},
\ \ \ \ 
R(i) = \sqrt{\frac{L_{o_x}(i)L_{o_y}(i)}{L_{g_x}(i)L_{g_x}(i)}}.
\label{eq:vcl_pr}
\end{equation}
The difference to the VCSL dataset metric~\cite{he2022vcsl} is the use of square-root, that we introduced to make the measure less strict, which we find to be more intuitive in several cases.

Figure \ref{fig:pr_explained} provides a visual explanation of the estimate of precision and recall for a toy case with 1 query video and 5 reference videos. 
%

%

\section{Challenge}

\label{sec:challenge}

The 2023 Video Similarity Challenge was held from December 2022 through April 2023.
This section describes the structure and rules of the challenge.

\subsection{Tracks}

The Challenge comprises two separate competition tracks,
namely the video copy detection track and the video copy localization track\footnote{The two tasks were referred to as ``descriptor track'' and ``matching track'' during the challenge}.%

\paragraph{\vcd track.}
Participants are required to provide a set of descriptors per video, which are used to perform detection with a standardized similarity-based detection approach.  
The descriptor dimensionality is restricted to be up to 512, and the total number of descriptors per video should not exceed an average rate of 1 descriptor per video second.
Detection is performed on an evaluation server by retrieving the most similar query and reference descriptor pairs globally, using inner-product similarity.
The maximum frame-wise similarity for each \textit{query-reference} pair is used as the detection confidence score for the whole videos.

\paragraph{\vcl track.}
This track also requires participants to provide their localization predictions.
The task is less restrictive on the approach: participants can use a variety of descriptor, detection, and localization techniques with a runtime complexity limitation.
On the server side, only evaluation of the predictions is performed.

\paragraph{Relationship to large-scale settings.}
Large-scale systems typically operate in stages, first retrieving candidate reference videos in a search step, then identifying copied segments between pairs in a localization step, since exhaustive localization is prohibitively expensive.
Our \vcd track models the first step of such a system, while our \vcl track models the end-to-end task of video copy detection and localization.
We impose compute limits on both tracks to constrain the challenge to techniques that are feasible at scale.

\subsection{Phases}

The challenge consists of two phases.
In Phase 1, the main competition phase, participants post predictions on the validation set, and scores are reported on a public leaderboard.
Participants have access to the train and validation sets in this phase.

Phase 2 is a short phase, designed to demonstrate that solutions generalize to a test set of new queries.
Phase 2 submissions must be based on code and data files submitted during Phase 1, demonstrating that solutions are not specific to the new query set.

\subsection{Rules}
\label{sec:rules}

\paragraph{Prediction independence.}
We follow \cite{douze2021isc} in requiring predictions to be independent.
This means that predictions for a \textit{query-reference} pair may not depend on other videos in the dataset.
This prohibits using techniques like PCA whitening trained on either the query or reference set being evaluated (but using the training set is acceptable).
This also requires retrieving the $k$ most similar descriptor pairs
jointly for all queries rather than the $k$ nearest neighbors to each query descriptor, as in ordinary $k$-nearest neighbor (KNN) search~\cite{douze2021isc}.

\paragraph{Validating submissions.}

Participants are required to produce code for their solutions to be eligible for prizes.
Solutions are submitted in the form of Docker containers including code and data files.
Submitted code is validated by extracting descriptors on a subset of dataset queries, performing detection and localization, and computing metrics.
This helps ensure that submitted code is sufficient to approximately reproduce results, and that submissions conform to challenge rules.
Code submission and validation is implemented by DrivenData, our challenge host platform.

\paragraph{Compute constraints.}
The validation step also allows checking the computational cost of submissions and enforcing a maximum resource constraint.
Validation ran in a container on reference hardware (6 vCPUs and 1 NVIDIA V100 GPU) so that the same compute constraints could be applied to all participants.
Submissions had to complete within a maximum time constraint of 10 seconds per query video, chosen to be $\approx 7\times$ slower than the SSCD baseline to accommodate more complex techniques.

\subsection{Baseline}

\newcommand{\igC}[1]{\frame{\includegraphics[height=2cm]{figs/low-complexity/pair_#1.png}}}
\begin{figure}
    \centering
    \begin{tabular}{cc}

    Q100097 & R109933 \\
    \igC{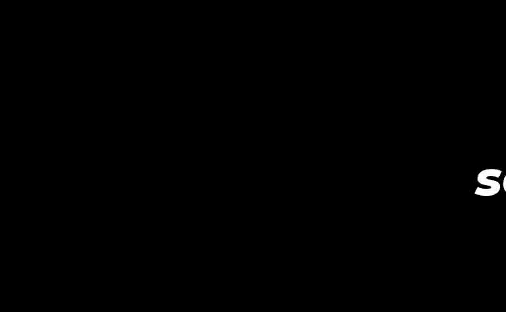} \imcreds{Q100097}{Ross Catrow, Gene Hunt} \imcreds{contd}{cammy8888, gosselinchristian} &
    \igC{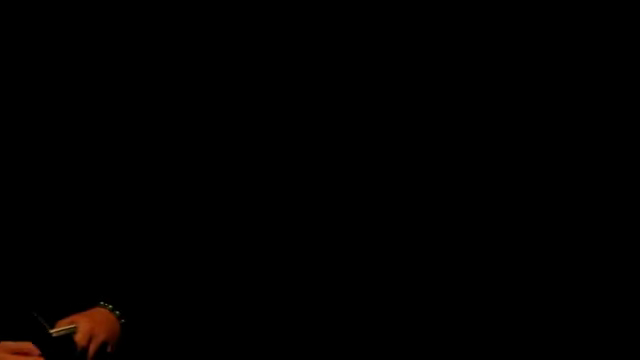}  \imcreds{R109933}{tedviens} \\

    Q103329 & R119106 \\
    \igC{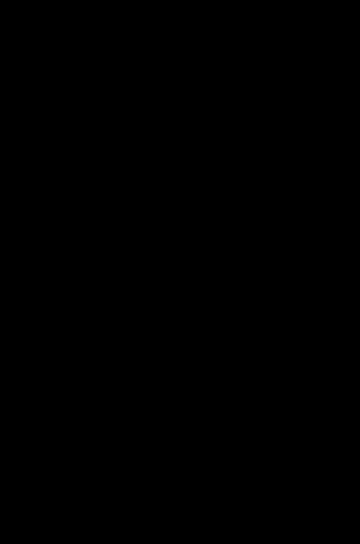} \imcreds{Q103329}{noii's} &
    \igC{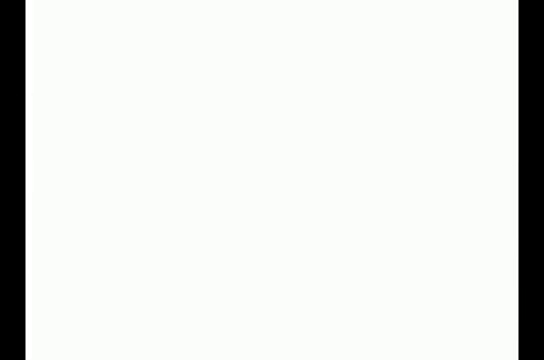} \imcreds{R119106}{PARTYGIRL///} \\

    Q100313 & R126160 \\
    \igC{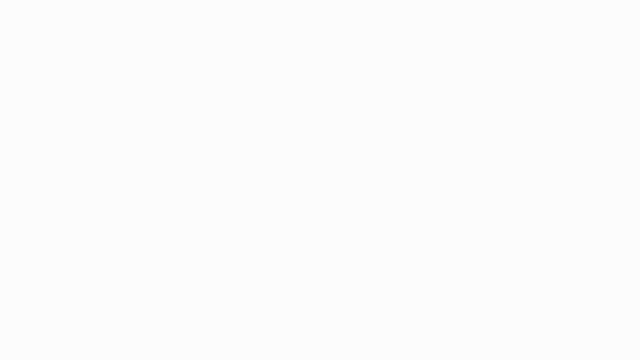} \imcreds{Q100313}{Lawrie Cate} &
    \igC{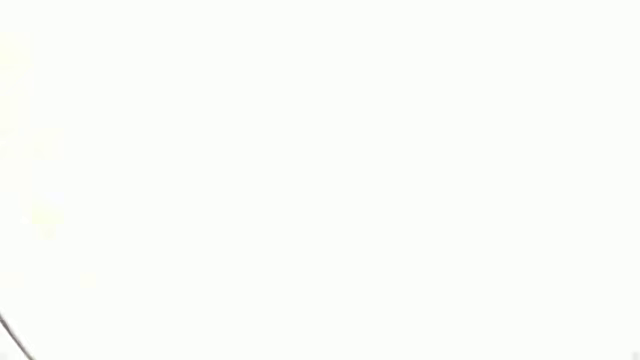} \imcreds{R126160}{YoTuT} \\
    \end{tabular}
    \caption{
        Examples (one per row) of low-information frames matches caused by high similarity when using L2 normalization.
    }
    \label{fig:low-complexity}
\end{figure}

We provide a baseline implementation to demonstrate a valid challenge submission for both tasks and to provide components participants can build on.

\paragraph{SSCD descriptor.}
Our baseline builds on SSCD~\cite{pizzi2022sscd}, a self-supervised image descriptor developed for copy detection.
It is a 512 dimensional descriptor from the ResNet50 SSCD model trained on the \DISC dataset with full augmentations.
We extract SSCD descriptors at one frame per second, after resizing the frame's short edge to 320 pixels and taking a center crop.

The original SSCD includes L2 normalization as a last step, which negatively impacts frames with low amount of visual content, \eg empty solid color frames as shown in Figure~\ref{fig:low-complexity}.
Those empty frames have descriptors with low norm, and therefore low inner-product similarity when L2 normalization is omitted. As a consequence, spurious empty frame matches are reduced.

We additionally include the score normalization process used in \DISC~\cite{douze2021isc,pizzi2022sscd}, where the descriptor similarity between two frames from the query and reference videos is refined by subtracting the similarity of the query video frame with its $k$-th most similar frame in the training set, multiplied by $\beta$. 
In particular we use $k=1$ and $\beta = 1.2$ for this normalization. 
To incorporate score normalization into the descriptor while retaining 512 dimensions and conforming to the challenge rules, the descriptor dimension with lowest variance is replaced with the score normalization term.

\begin{figure}
    \centering
    \includegraphics[width=\columnwidth]{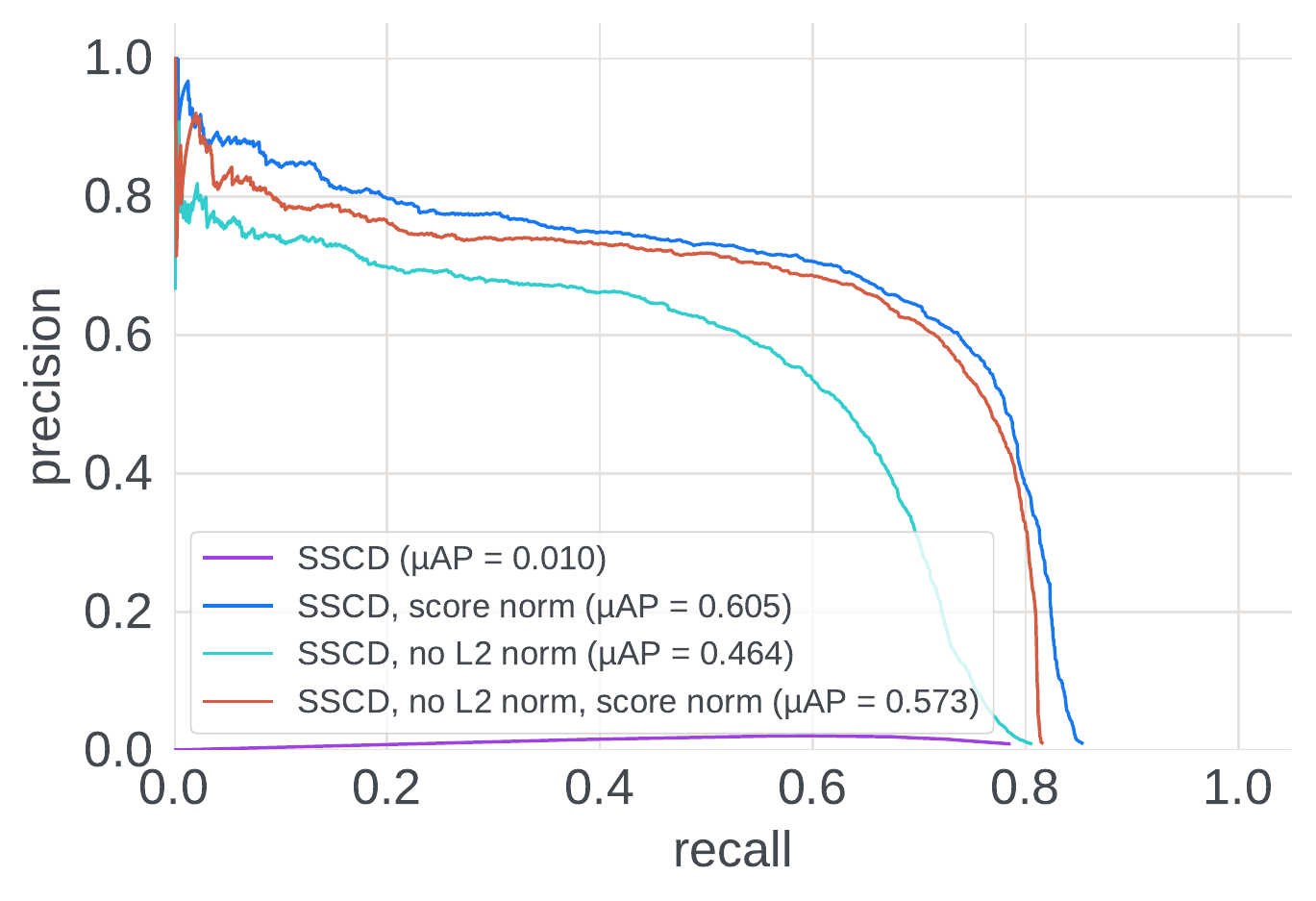}
    \caption{Test set VCD Track metrics for the SSCD baseline using various normalization strategies.}
    \label{fig:sscd_normalization}
\end{figure}

Results presented in Figure \ref{fig:sscd_normalization} show that the best results are achieved applying score normalization to L2 normalized descriptors, while without score normalization, the L2 normalized variant completely fails due to spurious empty frame matches.

\paragraph{Temporal network localization.}
We estimate temporal correspondences between candidate video pairs in a second step following descriptor search.
We compute a temporal similarity matrix $S = QR^\mathsf{T}$ as the product of query $Q$ and reference $R$ frame descriptors such that $S_{i,j}$ estimates the similarity between query frame $i$ and reference frame $j$.
This matrix is given as input to a temporal network~\cite{tan2009scalable} (see \ref{sec:related} Section), as implemented in \cite{he2022vcsl}, to estimate the localization in the form of temporal bounding boxes.
When using score normalization, a constant (0.5) is added to the similarity values because score normalized similarities can be negative and negative similarities are pruned from the temporal similarity graph and lead to poor results.
We assign each localized prediction a confidence score of the maximum $S_{i,j}$ value contained within the predicted bounding box.

\section{Results}

\label{sec:results}

\newcommand{\wechat}{\teamname{WeChat CV}\xspace}
\newcommand{\friend}{\teamname{Friendship First}\xspace}
\newcommand{\cvl}{\teamname{LINE CVL}\xspace}

In Phase 1, 48 teams submitted predictions for the \vcd track, and 18 teams submitted predictions for the \vcl track.
However fewer teams produced submissions that qualified for Phase 2, which requires code to run in our test environment and reproduce results on a subset of the dataset.
In Phase 2, 10 teams submitted \vcd track predictions, of which 5 outperformed the baseline, and 5 teams submitted \vcl track predictions, of which 4 outperformed the baseline.

\begin{table}[t]
    \centering
    \begin{tabular}{lrr}
        \toprule
                    team & \vcd track & \vcl track \\
        \midrule
            WeChat CV &       {\bf 87.2}\% &    {\bf 91.5}\% \\
        Friendship First &            85.1\% &          77.1\% \\
                     LINE CVL &            83.6\% &          70.4\% \\
        \midrule
                baseline &            60.5\% &          44.1\% \\
        \bottomrule
    \end{tabular}
    \caption{Top-ranked team and the achieved performance for the \vcd and \vcl tracks at phase 2.}
    \label{tab:team_metrics}
\end{table}

Table \ref{tab:team_metrics} shows the performance for the top-ranked teams across both challenge tracks.
In general a subset of teams that competed in the \vcd track competed in the \vcl track, and teams with strong \vcd track submissions also performed well on the \vcl track.
The finalists for both tracks were the same three teams in the same order.

Teams that participated in both tracks, with the exception of the top ranked team, have higher \vcd track scores than \vcl track.
While the \vcl track is less constrained, allowing a broader set of techniques to be used, it adds the additional requirement of localizing matches, and most solutions are penalized more for imperfect \vcl track localization than they make up in additional localization processing.

\newcommand{\teamname}[1]{\textsc{#1}}

\subsection{Methods}

We briefly analyze the methods used by the top-3 participants. 

\paragraph{Team \teamname{WeChat CV}~\cite{Wang2023dosomething,liu2023dosomething}.}
ViT~\cite{dosovitskiy2020image} is used as backbone and is trained with self-supervision as in SSCD~\cite{pizzi2022sscd}.

\textbf{\vcd track}: 
Frame descriptors are extracted with the trained backbone. Multiple descriptors are extracted for frames where multiple scenes are detected to be merged together. This detection process is performed based on simple image processing techniques. Score normalization is performed as in \DISC~\cite{douze2021isc}. 
An edited video detection process is filtering out the queries that are deemed not to have been transformed. 
This is the outcome of learning a binary classifier using CLIP~\cite{radford2021clip} for frame descriptors fed to RoBERTa~\cite{liu2019roberta} of edited and raw videos. 

\textbf{\vcl track}: 
Frame descriptors are extracted with the trained backbone and a temporal similarity matrix is obtained by comparing all pairs of frames between the two videos. 
This similarity matrix is processed in two different ways. 
(1) it is fed to a CNN-based binary classifier inferring whether the two videos are a copy of each other. 
Notably, despite the nature of the input similarity matrix, the CNN used is pre-trained on real images from ImageNet. 
This classifier constitutes a filtering step that reduces the amount of videos that need to be processed further. 
(2) The final step processes the similarity matrix to down-weigh the similarities between wrong frame correspondences and up-weigh the ones between correct correspondences. 
The strongest correspondences are processed by RANSAC to estimate the linear temporal transformation and identify inlier correspondences.

\paragraph{Team \teamname{Friendship First Competition Second}~\cite{wang2023friendshipfirst}.}

A CotNet~\cite{li2022contextual} is trained with self-supervision, in a metric learning manner, on synthetically generated image copies inspired by \DISC. 
This model is used as the teacher model to perform knowledge distillation on 7 different models, each based on a different network architecture. 
The video frames of the provided training set is used at this training stage. 
The trained models are used to form an ensemble model via descriptor aggregation.

\textbf{\vcd track}: Simple use of the ensemble model to extract frame descriptors.

 \textbf{\vcl track}: Temporal similarity networks~\cite{tan2009scalable} are applied on top of the temporal similarity matrix, which is estimated based on frame descriptors using the ensemble model.

\paragraph{Team \teamname{LINE CVL}~\cite{yokoo2023cvl}.}
The descriptor extraction is their model from the ISC2021 \vcd track~\cite{yokoo2021contrastive}, based on EfficientNetV2. 
To handle the case of videos concatenated together along the spatial dimension, each frame is split into fixed sized regions and one descriptor is extracted per region. 
These are stored as separate descriptors in the temporal direction with repeated timestamps. 
All descriptors of a video undergo a normalization process, while descriptors of consecutive frames are concatenated to capture temporal context and are further processed by PCA for dimensionality reduction. 
A classifier that detects edited videos so that unedited queries are filtered out and are not undergoing further processing.

\textbf{\vcd track}: Score normalization is performed as in \DISC~\cite{douze2021isc}

\textbf{\vcl track}: Temporal similarity networks~\cite{tan2009scalable} are applied on top of the temporal similarity matrix.

\paragraph{Common trends.}

All teams use a frame-level feature extractor as first processing step, \ie video-level descriptors are deemed unnecessary. 
Note that this was also the case for the baseline methods provided by the organizers. 
The concatenation and PCA method of team \teamname{LINE CVL} is a form of temporal descriptor though.

Two teams used an explicit detection of video transformations: \teamname{LINE CVL} and \teamname{WeChat CV} explicitly detect collages and process the individual collage elements separately. 
To a certain extent this exploits a bias in the dataset because in the ``rooms'' transform the videos are clearly separated, but the collage transformation is very common in real use cases. 

Both \teamname{LINE CVL} and \teamname{Friendship First} used the Temporal Network method~\cite{tan2009scalable} that was provided in the baseline method to aggregate frame-level matches into video-level matches. 
It constructs a graph where nodes are frames and weighted edges are between matching frames. 
The video matching is then cast as a path finding operation in the graph. 

The teams \teamname{WeChat CV} and \teamname{LINE CVL} use the score normalization procedure based on a ``background distribution'' of negative images~\cite{douze2021isc}. 

Teams \teamname{WeChat CV} and \teamname{LINE CVL} use a classifier to detect edited query videos. This approach exploits dataset bias given the knowledge that all copied queries include temporal edits which are easy to detect by such an approach. Nevertheless, it is not a realistic real-world solution that would otherwise fail in the presence of harder to detect transformations, \eg small spatial and temporal crops, which are challenging for video copy detection and localization. 

\subsection{Analysis}

\begin{figure}[t]
    \centering
    \includegraphics[width=1.00\columnwidth]{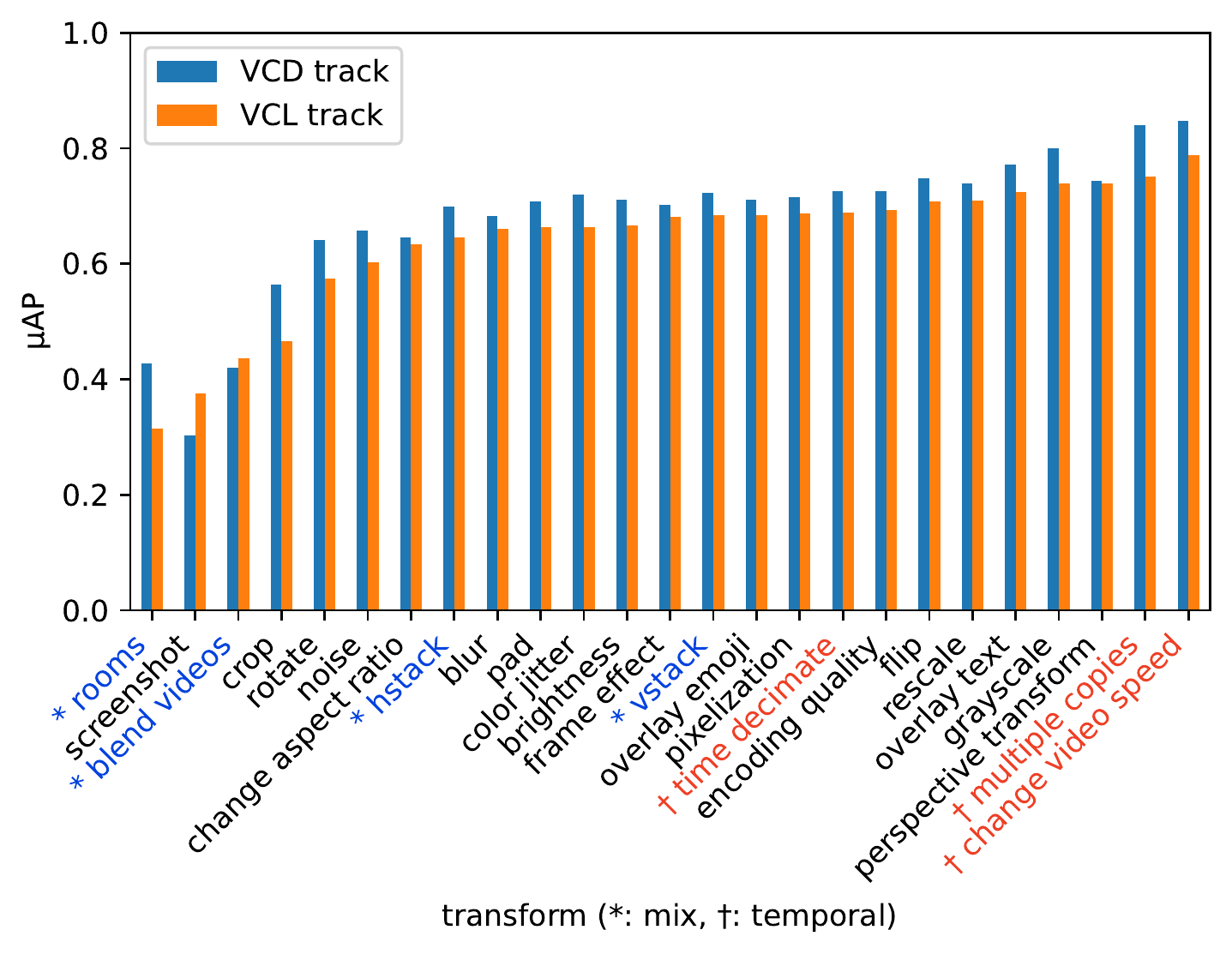}
    \caption{Mean test set accuracy of top three teams by transform.
    }
    \label{fig:transform_metrics}
\end{figure}

\begin{figure}[t]
    \vspace{-20pt}
    \centering
    \includegraphics[width=0.95\columnwidth]{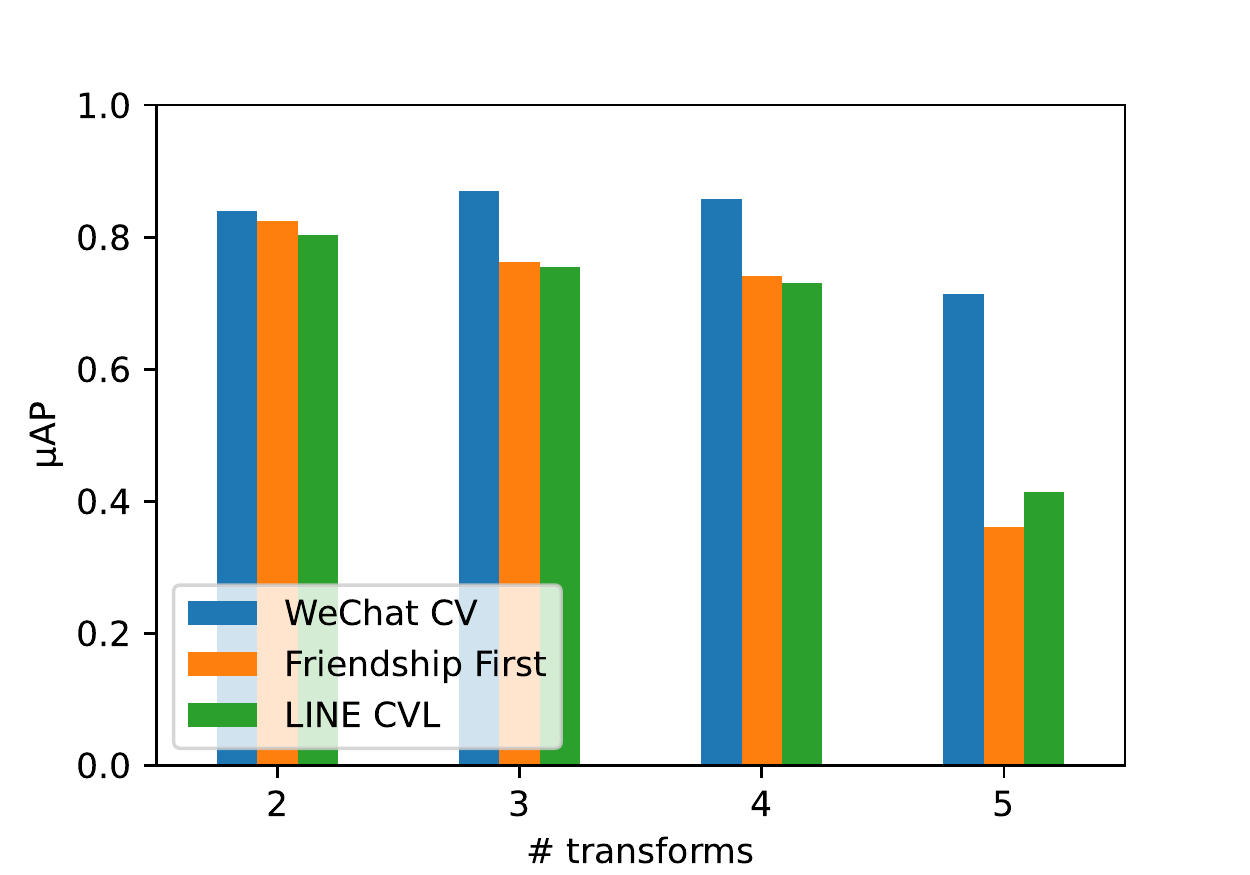}
    \includegraphics[width=0.95\columnwidth]{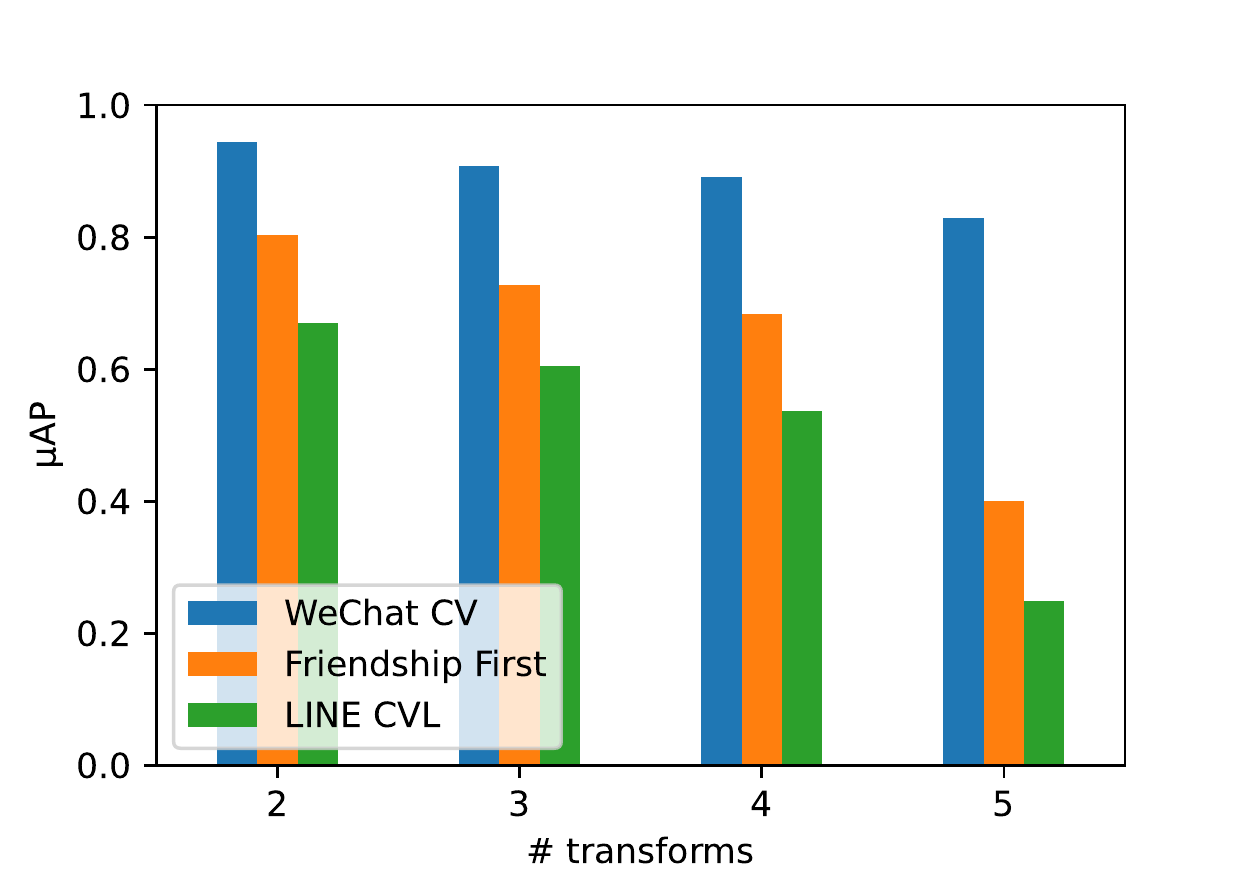}
    \vspace{-10pt}
    \caption{Mean test set accuracy of top three teams by number of transforms applied. Above: \vcd track; below: \vcl track.
    }
    \label{fig:num_transforms}
\end{figure}

We analyze performance of the top three submissions for each track to extract general trends on this dataset.
For each setting, we evaluate using a subset of queries with copied segments, and  all distractor queries.

\paragraph{Analysis per transformation.}
To compare the difficulty of each transformation in the dataset, we compute metrics on query subsets that include that transformation.
Figure \ref{fig:transform_metrics} shows performance per transformation for both tracks.

We find that transformation difficulty for both tracks are well correlated.
Temporal transformations such as speed change and inserting multiple copies are among the easiest transformations, especially for the \vcd track, since all submissions build on frame-based descriptors.
The most challenging transformations are those that mix frames from multiple videos (marked with * in Figure \ref{fig:transform_metrics}), and geometric transforms that add or remove image surface such as overlay onto screenshot and crop.

\paragraph{Analysis per number of transforms.}
We perform a similar analysis of accuracy per number of transformations applied in Figure \ref{fig:num_transforms}.
We find that videos with more transformations applied are more difficult, especially for \vcl track. Notably, teams with video editing detection are more robust to the number of transformations, i.e., \wechat achieves better \uAP when 3 and 4 instead of 2 transformations have been applied, and \cvl outperforms \friend on 5 transformations.

\paragraph{Effect of distractor queries.}
We compute metrics on a subset of the dataset, excluding distractor queries.
We find that the average scores for the finalist teams improve by $4\%$ on average for both tracks, although \teamname{WeChat CV}'s scores change by $ <1\%$, which may indicate the effectiveness of edit detection on this dataset.
This has the effect of narrowing the range of scores in the VCL track, and changing the order of finalists in the VCD track.

\begin{figure}[t]
    \centering
    \includegraphics[width=.9\columnwidth]{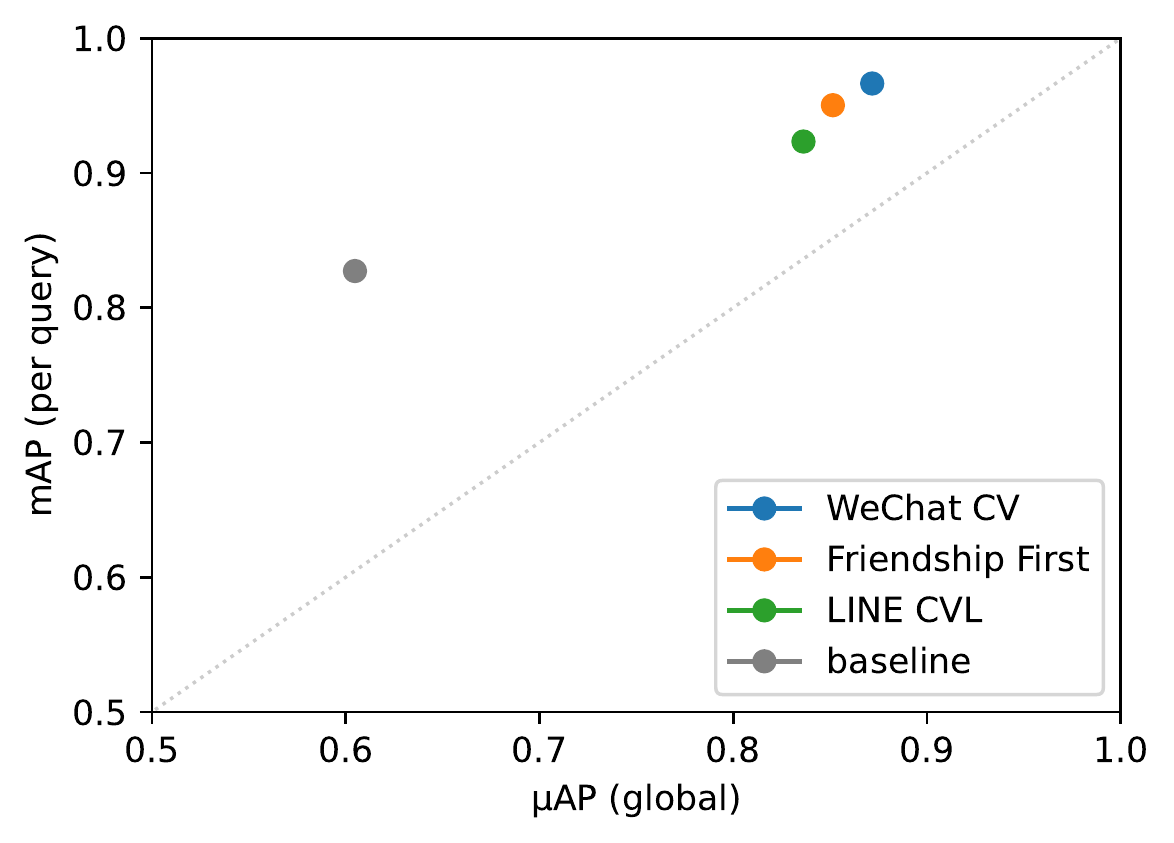}
    \vspace{-10pt}
    \caption{Comparing \uAP and mAP metrics for top VCD track submissions.}
    \label{fig:map-comparison}
\end{figure}

\begin{figure}[h!]
    \centering
        \includegraphics[width=.9\columnwidth]{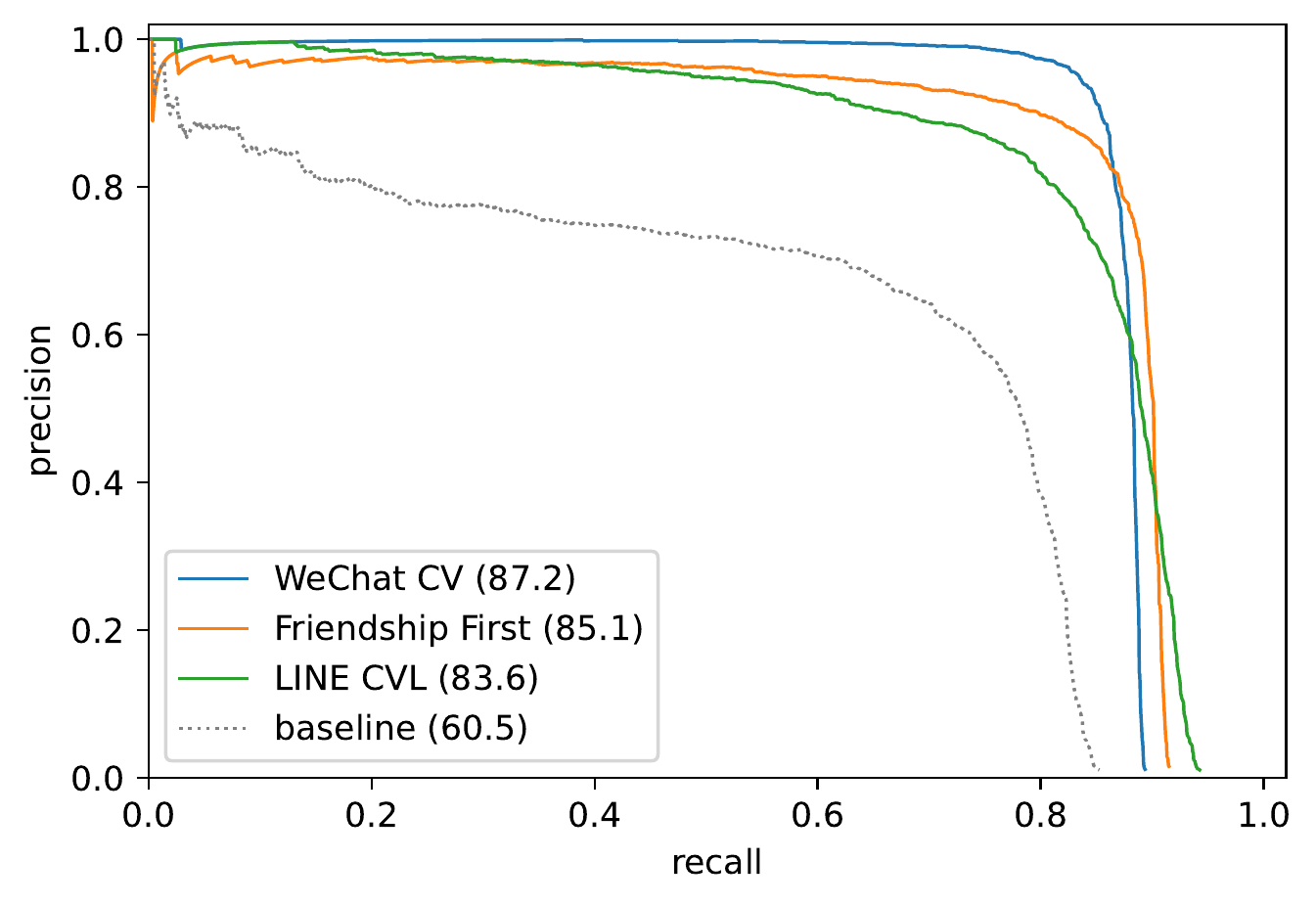}
        \includegraphics[width=.9\columnwidth]{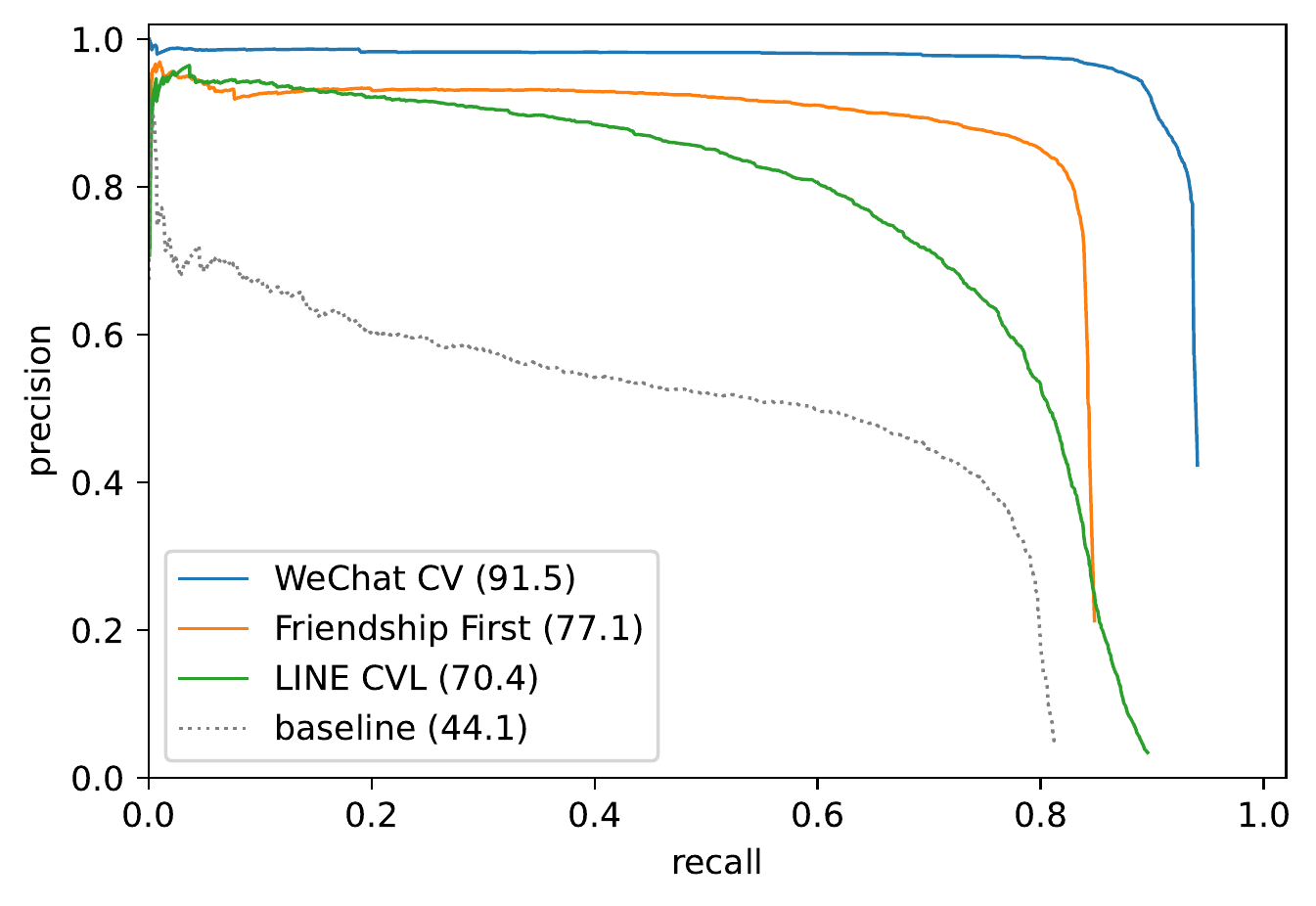}
    \vspace{-20pt}
    \caption{Precision-recall curves for the VCD track (top) and VCL track (bottom). Precision and recall are defined conventionally for the VCD track, and as in Equation \ref{eq:vcl_pr} for the VCL track. }
    \label{fig:pr-plots}
\end{figure}

\paragraph{Comparison with mAP.}

We evaluate VCD track submissions with mAP, and compare how calibration across queries affects scores in Figure \ref{fig:map-comparison}.
This has the additional effect of excluding distractor queries, since mAP is defined for queries with ground truth matches.
Submissions score $+9.4\%$ higher with mAP than \uAP, and this trend is similar for the top teams (within $1\%$).
Top submissions are notably better calibrated than the baseline, which has a $22.2\%$ gap between mAP and \uAP metrics.

\begin{figure*}[t]
    \centering
    \includegraphics[width=0.32\linewidth]{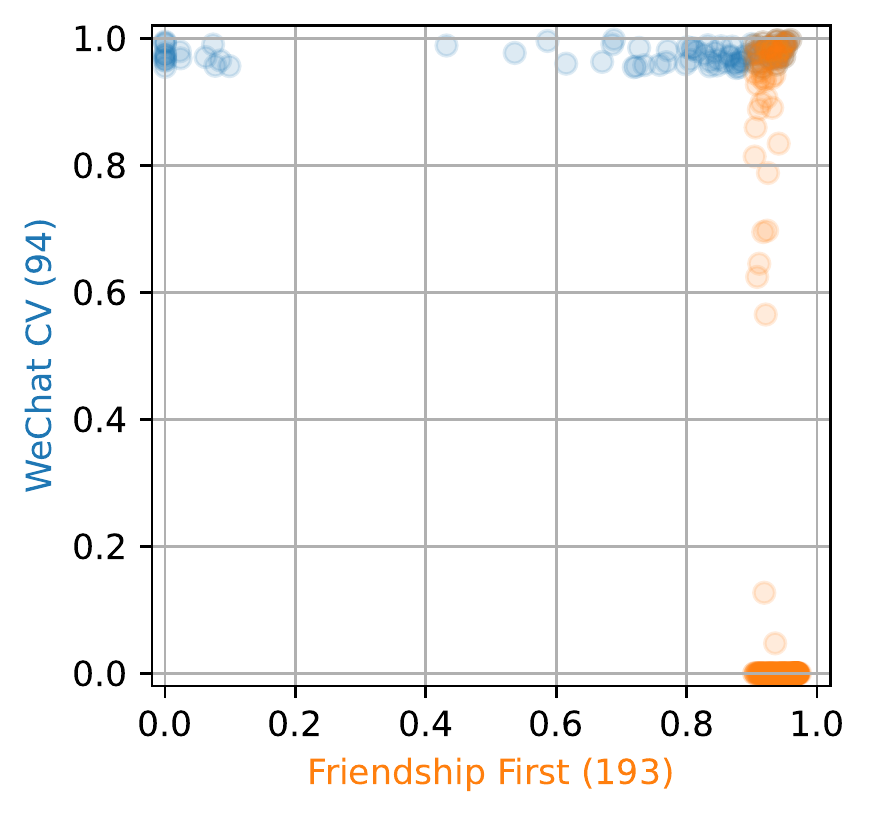}
    \includegraphics[width=0.32\linewidth]{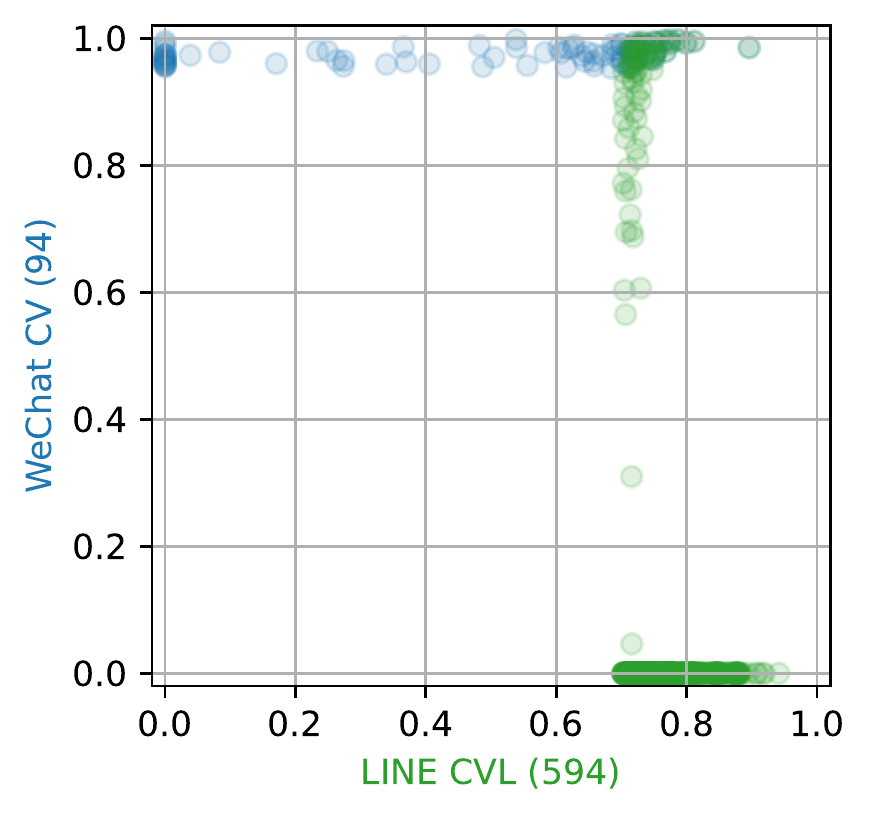}
    \includegraphics[width=0.32\linewidth]{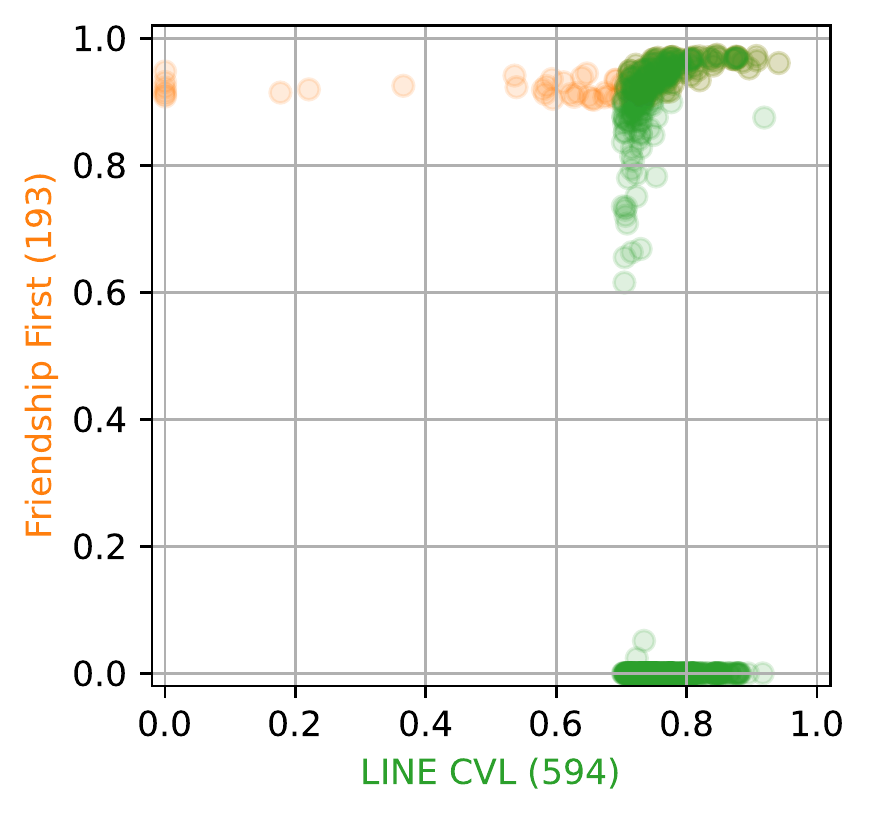}
    \caption{How are the top-ranked negatives of one method ranked for another one? 
    Precision values shown per method for sets of hard negatives according to each of the methods. Each point corresponds to a hard negative pair ranked among the top-2K ranks on the \vcd track according to each method and is colored according to the method it is hard negative for.
    }
    \label{fig:precision_scatters}
\end{figure*}

\paragraph{Precision-recall curves.}

Figure \ref{fig:pr-plots} visualizes precision and recall for the top submissions for both tracks.
In the \vcd track, all participants find a similar amount of recall in total, and precision accounts for most of the difference in score.
\wechat is the only team that achieves a higher score on the \vcl track than on the \vcd track.
This demonstrates very accurate localization, and shows a benefit of the techniques the team introduced to reduce spurious detection.

\paragraph{Method comparison via hard negatives.}

For each method in the \vcd track, we find the set negatives that are ranked among the top 2000 pairs. This results in three sets of pairs, one per method. Then, for each method, we compute the precision $P(i)$ at the rank $i$ at which each of these pairs, from all three lists, is ranked. 
A comparison between every combination of two methods for all 3 teams is shown in Figure \ref{fig:precision_scatters}. 
There are several common hard negative pairs for the compareds methods, \ie points in the upper-right corner close to [1, 1]. However, plenty of points are hard negative for one method, \ie high precision score, but not for the other, \ie low precision score. This highlights the large amount of complementarity between the methods, so one method can better handle some hard negatives than the other.

\vspace{-2mm}
\section{Conclusion}

The DVSC2023 aims at becoming a reference dataset for video copy detection, a task of high practical importance for content sharing platforms. 
It comes with a thoughtful design, clean legal terms (in contrast with some datasets that come in the form of %
URLs that users need to scrape themselves), and solid baseline methods.
Participants used frame-based feature extraction with various strong image models. 
These features are then combined to generate video-level matches. 
The main outcome from the competition is that these state-of-the-art methods are very capable of recognizing videos despite extensive transformation.
We will make sure that the dataset remains available during the coming years to foster research in this domain.

\bibliographystyle{plain}
\bibliography{biblio}

\end{document}